\documentclass{article}
\usepackage[T1]{fontenc}
\usepackage[utf8]{inputenc}
\usepackage[preprint]{colm2025_conference}
\usepackage[letterpaper,margin=1.25in]{geometry}
\usepackage{graphicx}
\usepackage{capt-of}
\usepackage{wrapfig}
\usepackage{needspace}
\usepackage{amsmath,amssymb}
\usepackage{array,booktabs,tabularx,colortbl}
\usepackage{placeins}
\usepackage{hyperref,eucal}
\definecolor{mblue}{HTML}{0064E0}
\hypersetup{colorlinks=true,linkcolor=mblue,citecolor=mblue,urlcolor=mblue}
\fancypagestyle{firstpagefancy}{
  \fancyhf{}
  \fancyhead[L]{\raisebox{-0.15\height}{\includegraphics[height=1.2\baselineskip]{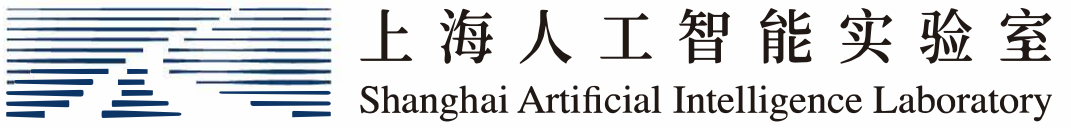}}}
  
  \fancyfoot[C]{\thepage}
}
\graphicspath{{figures/}}
\input{nickname}
\newcommand{\stopgrad}{\operatorname{sg}}
\title{\nickname: A Foundation Model for 4D Scene Geometry and Motion}
\hypersetup{pdftitle={\nicknametext: A Foundation Model for 4D Scene Geometry and Motion},
  pdfauthor={Yifan Wang, Weiguang Zhao, Ni Yang, Haoyu Guo, Junting Dong, Weinan Zhang, Chunhua Shen}}
\author{\normalfont\small Yifan Wang\textsuperscript{1,2}, Weiguang Zhao\textsuperscript{1}, Ni Yang\textsuperscript{1}, Haoyu Guo\textsuperscript{1,}\thanks{Corresponding authors.},~
  Junting Dong\textsuperscript{1,*}, Weinan Zhang\textsuperscript{1,2}, Chunhua Shen\textsuperscript{1,3,*}\\[4pt]
  {\normalfont\small\textsuperscript{1}Shanghai Artificial Intelligence Laboratory\quad
    \textsuperscript{2}Shanghai Jiao Tong University\quad\textsuperscript{3}Zhejiang University}\\[3pt]
  {\normalfont\small
    \href{https://github.com/yyfz/Pi4}{Code}\quad\textbar\quad
    \href{https://yyfz.github.io/pi4}{Project Page}\quad\textbar\quad
    \href{https://huggingface.co/spaces/yyfz233/Pi4}{Hugging Face Demo}}}
\begin{document}
\begingroup
\setlength{\tabcolsep}{0pt}
\maketitle
\endgroup
\begin{figure}[!ht]
  \centering
  \includegraphics[width=\linewidth]{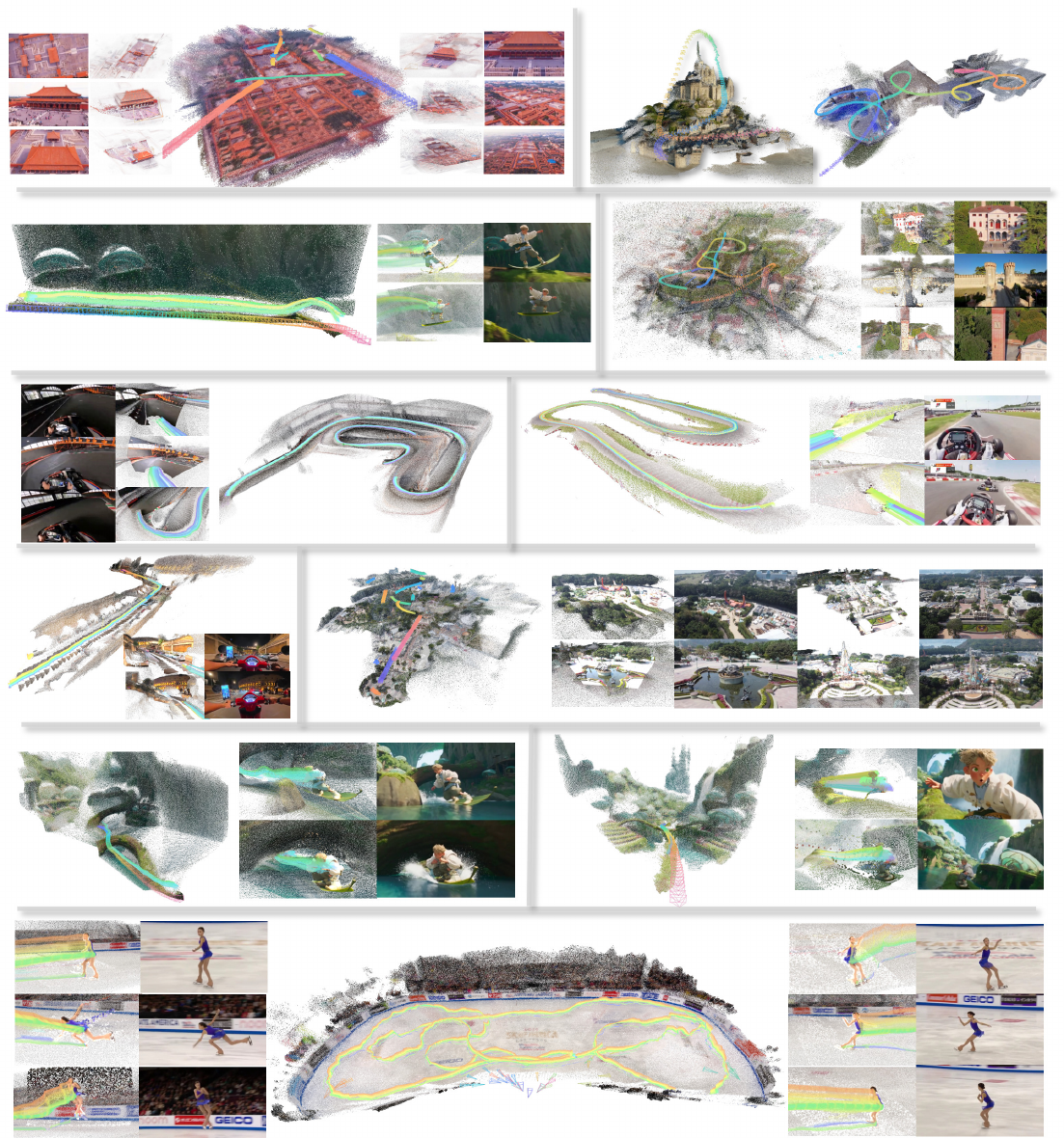}
  \caption{\textbf{Geometry and motion in a single forward pass.} All reconstruction and tracking results are obtained in a \textbf{single network forward pass} per scene, including cases with \textbf{over 1{,}000 input images}.}
  \label{fig:teaser}
\end{figure}
\clearpage
\begin{abstract}
\nickname{} is a unified foundation model for 4D scene perception. From diverse, uncalibrated RGB observations in the wild---a single image, an unordered collection, or a video sequence---it recovers dense geometry, camera poses, and metric scale, together with 2D/3D point tracks and moving-object masks for multi-image inputs. It couples a geometry branch with a geometry-conditioned motion branch, which predicts image-space correspondences that are lifted into 3D using reconstructed target-view geometry. Across diverse benchmarks, \nickname{} achieves the best overall average rank in geometry evaluation, alongside competitive performance in single- and multiview 2D/3D point tracking and moving-object segmentation. With a shared backbone, the full geometry-and-motion system contains approximately 0.5B parameters---less than half the size of most baselines. It produces joint reconstructions and tracks in one forward pass, including large-scale scenes with thousands of input images despite training with contexts of at most 64 views. We further present the data and training recipe underlying these capabilities, together with empirical studies of key design choices.
Code and models are available at \href{https://github.com/yyfz/Pi4}{Pi4}.
\end{abstract}

\section{Introduction}
\label{sec:introduction}

Feed-forward models have established a powerful paradigm for reconstructing scenes directly from RGB observations~\citep{wang2024dust3r,wang2025vggt,wang2025pi3}. More recently, this paradigm has been extended beyond scene reconstruction to encompass motion recovery~\citep{lu2026track4world,fradlin2026arrow}. A key question, however, is how well these capabilities generalize to in-the-wild observations, where scene content, camera configurations, temporal continuity, and the number of observations are unconstrained. Maintaining high-quality geometric reconstruction while robustly recovering motion from such diverse inputs remains a significant practical challenge.

\begin{wrapfigure}{r}{0.4167\textwidth}
  \vspace{-10pt}
  \centering
  \includegraphics[width=\linewidth]{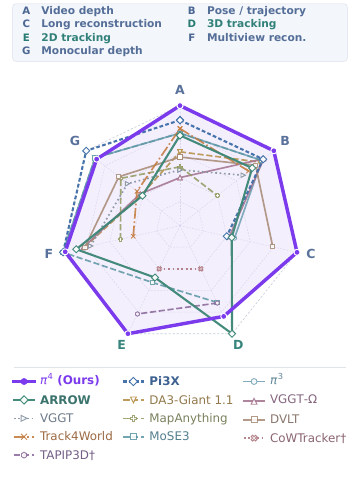}
  \caption{\textbf{Geometry and motion across seven task groups.} Task-wise normalized aggregate scores. Gaps denote missing or incomplete results. $\dagger$: VGGT-$\Omega$ geometry.}
  \label{fig:performance-radar}
\end{wrapfigure}

To address this challenge, we present \nickname{}, a unified foundation model for 4D scene perception from RGB observations. It handles single images, unordered image collections, and videos, recovering dense geometry, camera poses, and metric scale, together with 2D and 3D point tracks and moving-object masks for multi-image inputs. Across the geometry benchmarks in Table~\ref{tab:geometry-results}, \nickname{} achieves the best overall average rank among the compared methods, while delivering competitive performance in single- and multiview 2D/3D tracking and moving-object segmentation (Figure~\ref{fig:performance-radar}). Despite this broad functionality, the model employs a shared backbone across these capabilities and contains approximately 0.5B parameters in total---less than half the size of most baselines. Beyond benchmark results, Figure~\ref{fig:teaser} showcases joint geometry and motion reconstruction across diverse in-the-wild observations, including aerial footage, casual handheld recordings, driving and sports scenes, and stylized content. These reconstruction and tracking results are all produced in a single forward pass per scene. This combination of accuracy, efficiency, and generality is enabled by coordinated choices in model architecture, training, and data.

At the architectural level, the geometry network of \nickname{} retains the permutation-equivariant formulation of $\pi^3$~\citep{wang2025pi3}, treating all input views symmetrically without designating a reference view. Both the geometry and motion branches employ looped Transformer decoding~\citep{burzio2026dejaview}, which we adopt for its strong cross-benchmark performance and parameter efficiency. We choose loop weight sharing partly to reduce memory usage, freeing capacity for larger within-sequence contexts. Training stochastically samples view counts from single images, 2--24-view sets, and 32/48/64-view sets, with short contexts dominating the mixture. A sparse 3D refiner~\citep{kong2026moge3} further enhances local surface structure and geometric detail.

To incorporate motion estimation without compromising reconstruction, we freeze the geometry network and train a separate geometry-conditioned motion decoder. Direct 3D tracking couples correspondence estimation with geometry, camera placement, and scene scale in the regression target. We instead keep motion prediction in image space and delegate 3D reasoning to the reconstruction branch, composing the two analytically.

\Needspace{30\baselineskip}

This yields a simple correspondence objective that is independent of scene scale, while placing every visible 3D track point on the reconstructed target-view surface. Geometry features also provide useful context for correspondence prediction.

Complementing these architectural and modeling choices, our data construction pipeline addresses gaps in scene and camera coverage, as well as the difficulty of obtaining geometric supervision for foreground-dominated videos. We render scene-level assets from Funes~\citep{funesworld} and Sketchfab~\citep{sketchfab}, covering architectural environments and other large-scale scenes rather than isolated objects alone. Coverage-aware view selection and connected view sampling balance broad scene coverage with local cross-view matchability, while complementary natural and counterbias camera configurations expand the range of roll angles and focal lengths encountered during training. We further apply field-of-view and lens augmentation to diversify image-to-ray mappings beyond those induced by conventional pinhole or rectified imagery. For Internet videos dominated by moving foregrounds, we infer background geometry following foreground suppression and video inpainting. We retain only points corresponding to originally visible background pixels and pair the resulting pseudo-labels with the original RGB frames. \emph{GameRealm}~\citep{gamerealm}, developed jointly with collaborators, further provides complementary engine-native geometry and persistent point trajectories as supervision for dynamic scenes.

Together, these architectural, data, and training choices enable coherent joint reconstruction and tracking from thousands of images in large-scale scenes despite a maximum training context of 64 views (Figure~\ref{fig:teaser}). We further examine the architecture and training strategy of \nickname{} through targeted ablations and qualitative comparisons, highlighting the empirical findings and practical trade-offs that motivate these choices.

\section{Related Work}
\label{sec:related-work}

\subsection{Visual Geometry Reconstruction}

Classical structure-from-motion (SfM) estimates cameras and sparse structure through image matching and bundle adjustment, while multi-view stereo (MVS) reconstructs dense surfaces from calibrated views. COLMAP exemplifies this modular pipeline~\citep{schoenberger2016sfm,schoenberger2016mvs}. Learned MVS methods such as MVSNet~\citep{yao2018mvsnet} learn cost-volume regularization while retaining camera calibration as input. More recently, DUSt3R~\citep{wang2024dust3r} predicts pairwise pointmaps without supplied cameras, and MASt3R~\citep{leroy2024mast3r} combines this 3D representation with learned matching descriptors, bridging reconstruction and correspondence estimation.

Feed-forward multi-view models increasingly estimate geometry and cameras jointly. VGGT~\citep{wang2025vggt} predicts cameras, depth, pointmaps, and tracks from image sets, while $\pi^3$ and its released Pi3X variant~\citep{wang2025pi3} use permutation-equivariant modeling without a privileged input reference. MapAnything~\citep{keetha2026mapanything} targets metric reconstruction with flexible geometric inputs, whereas Depth Anything 3~\citep{chen2025depthanything3} uses depth and ray prediction across arbitrary views. VGGT-$\Omega$~\citep{wang2026vggtomega} incorporates matching supervision, and DVLT~\citep{burzio2026dejaview} reuses Transformer blocks through looped decoding. Monocular models provide complementary components: MoGe2~\citep{wang2025moge2} separates metric-scale prediction from relative geometry, and MoGe3~\citep{kong2026moge3} refines surface detail with sparse volumetric processing. Together, these methods provide complementary designs for joint camera--geometry estimation, reference-free reconstruction, metric prediction, and iterative decoding.

\subsection{Image Correspondence and Point Tracking}

Classical image tracking estimates motion through local appearance alignment, as in Lucas--Kanade~\citep{lucas1981iterative}. Modern optical-flow models such as RAFT~\citep{teed2020raft} learn dense two-frame correspondence, while TAPIR~\citep{doersch2023tapir} and CoTracker3~\citep{karaev2024cotracker3} follow queried points over videos and estimate visibility. CoWTracker~\citep{lai2026cowtracker} replaces correlation-volume lookup with iterative feature warping. Optical flow and query-based tracking thus provide dense or sparse image-space correspondences, typically within ordered video sequences.

\subsection{Dynamic Scene Reconstruction and 4D Perception}

Dynamic scene reconstruction extends geometry estimation to scenes with moving and deforming content. Shape of Motion~\citep{wang2025shapeofmotion} reconstructs 4D scenes from monocular videos using compact $\mathrm{SE}(3)$ motion bases and data-driven geometric and tracking priors. MonST3R~\citep{zhang2025monst3r} instead adapts pointmap reconstruction to dynamic videos by estimating per-timestep geometry without an explicit motion representation. These approaches connect scene geometry with temporal variation, but differ in how motion is represented and how reconstruction and tracking are coupled.

Recent methods further extend tracking to 3D and connect it to scene reconstruction. D4RT~\citep{zhang2026d4rt} and ARROW~\citep{fradlin2026arrow} use query-based decoding to recover point positions across time and reference cameras; ARROW additionally supports unordered and multiview observations. MoSE3~\citep{cheng2026mose3} models dense world-space motion through per-pixel $\mathrm{SE}(3)$ transformations, while Track4World~\citep{lu2026track4world} combines dense correspondences, geometry, and camera estimation for world-centric tracking.

Moving-object segmentation provides a complementary object-level characterization of scene dynamics by identifying regions that undergo motion~\citep{tokmakov2019moving,xie2026gmos}. Across these directions, methods differ in their supported input configurations, output representations, and the degree to which geometry and motion are modeled jointly.

\section{Method}
\label{sec:method}

\begin{figure}[!t]
  \centering
  \includegraphics[width=\linewidth]{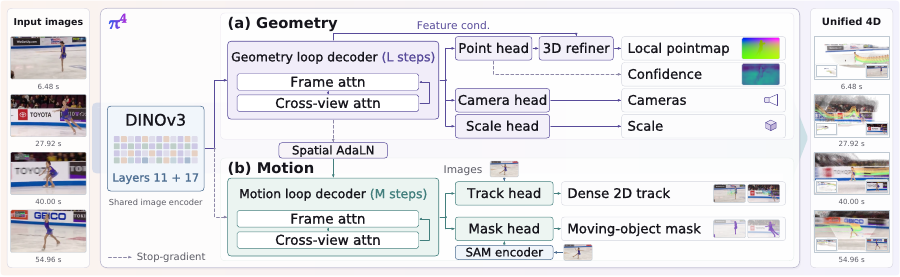}
  \caption{\textbf{\nickname{} overview.} Shared DINOv3 features feed geometry and motion loop decoders for reconstruction, dense tracking, and moving-object segmentation. A sparse 3D refiner improves local geometry, while detached geometry features condition motion through spatial adaptive layer normalization (AdaLN).}
  \label{fig:pipeline}
\end{figure}

\nickname{} combines geometric reconstruction and motion estimation within a single feed-forward model (Figure~\ref{fig:pipeline}). A shared image encoder provides features to two task-specific looped Transformer decoders. The geometry branch predicts dense pointmaps, camera poses, and metric scale; a sparse 3D refiner then improves the predicted pointmaps. Conditioned on geometry features, the motion branch predicts image-space point tracks and moving-object masks. The predicted tracks are lifted to 3D using the reconstructed target-view geometry.

\subsection{Task Formulation and Output Representations}
\label{sec:method-task}

Given $N$ RGB images $\mathcal{I}=\{\mathbf{I}_i\}_{i=1}^{N}$, with $\mathbf{I}_i\in\mathbb{R}^{H\times W\times 3}$, \nickname{} predicts scene geometry and, for multi-image inputs, point tracks and moving-object masks. Camera intrinsics and extrinsics are not required as inputs, and the images need not follow a chronological ordering.

The geometric outputs comprise a dense local pointmap $\mathbf{P}_i\in\mathbb{R}^{H\times W\times 3}$, a camera-to-world pose $\mathbf{T}_i=(\mathbf{R}_i,\mathbf{t}_i)\in\mathrm{SE}(3)$, and a confidence map $c_i(\mathbf{u})\in[0,1]$ for each image, together with a positive scale $m$ shared across the inputs. Points are expressed in their respective camera coordinates; the camera poses relate these local reconstructions through a common scene frame, and $m$ converts the reconstruction to metric units.

For motion tracking, a query view $q$ specifies the points to track. The model predicts dense image-space correspondences $\mathbf{U}_{q\to i}:\Omega_q\to\mathbb{R}^2$, together with visibility and tracking confidence. Here $\Omega_q$ is the query pixel grid, and $\mathbf{U}_{q\to i}(\mathbf{u})$ gives the corresponding location in image $i$. World-space tracks $\mathbf{Y}_{q\to i}(\mathbf{u})$ combine these correspondences with the reconstructed geometry. The model also predicts a moving-object mask for each image, independently of the query selection.

\subsection{Shared Encoding and Looped Decoding}
\label{sec:method-loop}

\paragraph{Image encoder.}
A shared DINOv3 encoder~\citep{simeoni2025dinov3} extracts per-image patch features for both branches. We combine intermediate and late-layer encoder features, project them into the decoder feature space, and append shared learnable register tokens. We examine the choice of encoder layers in Section~\ref{sec:feature-lessons}.

\paragraph{Looped decoders.}
Both the geometry and motion decoders alternate frame and cross-view attention~\citep{wang2025vggt,wang2025pi3}. Following D\'ej\`a View~\citep{burzio2026dejaview}, each decoder repeatedly applies the same pair of blocks, sharing their weights across its loop steps. The two decoders have independent parameters; weight sharing occurs within each branch, not between branches. Starting from encoded tokens $\mathbf{Z}^{0}$, the geometry decoder applies
\begin{equation}
  \mathbf{Z}^{\ell+1}
  =\mathcal{B}_{\mathrm{cross}}\!\left(
    \mathcal{B}_{\mathrm{frame}}\!\left(\mathbf{Z}^{\ell};\tau_\ell\right);
    \tau_\ell\right),
  \qquad \ell=0,\ldots,L-1,
  \label{eq:method-loop}
\end{equation}
where $\mathcal{B}_{\mathrm{frame}}$ is a Transformer block that processes each image independently, and $\mathcal{B}_{\mathrm{cross}}$ is a Transformer block that aggregates information across input views. The conditioning variable $\tau_\ell$ encodes the normalized current and next loop positions. The geometry loop count $L$ is sampled between 8 and 16 during training, whereas motion decoding uses a fixed $M=8$. Evaluation uses $L=16$ and $M=8$. Task heads read only the final decoder state.

\paragraph{Geometry-conditioned motion decoding.}
The motion decoder starts from image tokens and adopts the same frame- and cross-view-attention block design as the geometry decoder, but with independent parameters. It follows the same looped attention structure, with additional geometry conditioning: spatial adaptive layer normalization (AdaLN) conditions its attention and feed-forward sublayers on detached geometry features at the same view and patch, together with loop progress. A final within-image readout supplies the tracking and moving-object mask heads. The motion decoder uses no temporal positional embeddings.

\subsection{Geometric Reconstruction}
\label{sec:method-geometry}

\paragraph{Reference-free geometry.}
\nickname{} retains the permutation-equivariant geometry formulation of $\pi^3$~\citep{wang2025pi3}: reordering the images only reorders the corresponding geometry and camera predictions. The geometry decoder uses no view-order or reference-view embeddings. Rather than prescribing a reference camera or world coordinate frame, we use relative camera supervision to let the network learn a consistent common frame from the relations between views.

\paragraph{Point and camera prediction.}
For a pixel $\mathbf{u}$, we parameterize its local point by a positive depth $z_i(\mathbf{u})$ and a camera-local ray $\mathbf{r}_i(\mathbf{u})$:
\begin{equation}
  \mathbf{r}_i(\mathbf{u})=
  \begin{bmatrix}r_i^x(\mathbf{u}) & r_i^y(\mathbf{u}) & 1\end{bmatrix}^{\!\top},
  \qquad
  \mathbf{P}_i^{(0)}(\mathbf{u})=z_i(\mathbf{u})\mathbf{r}_i(\mathbf{u}).
  \label{eq:method-local-points}
\end{equation}
The ray is not unit-normalized, and $z_i$ denotes camera-axis depth. A within-image readout feeds a convolutional upsampling head that predicts $(r_i^x,r_i^y,\log z_i)$, yielding the coarse pointmap $\mathbf{P}_i^{(0)}$. A shallow camera decoder regresses per-image poses, and a context-attending decoder predicts one log-scale $\ell_m$ per input set, giving $m=\exp\ell_m$.

\paragraph{Sparse geometry refinement.}
\label{sec:method-refiner}
Following MoGe3~\citep{kong2026moge3}, a sparse 3D U-Net refines each view's geometry in ray/log-depth coordinates, conditioned at its bottleneck on projected geometry-decoder features. Starting from the coarse depth $z_i^{(0)}$, it updates log-depth while retaining the predicted rays:
\begin{equation}
  \log z_i^{(s+1)}=\log z_i^{(s)}+\Delta_i^{(s)},
  \qquad
  \mathbf{P}_i^{(s+1)}=z_i^{(s+1)}\mathbf{r}_i,
  \quad s=0,\ldots,S-1.
  \label{eq:method-refine}
\end{equation}
The final pointmap is $\mathbf{P}_i=\mathbf{P}_i^{(S)}$.

\paragraph{Metric reconstruction and confidence.}
The predicted camera poses and scale place the local geometry in a common metric frame:
\begin{equation}
  \mathbf{X}_i^{\mathrm{metric}}(\mathbf{u})
  =m\bigl(\mathbf{R}_i\mathbf{P}_i(\mathbf{u})+\mathbf{t}_i\bigr).
  \label{eq:method-world-points}
\end{equation}
No ground-truth alignment is required at inference. A convolutional confidence branch shares early local-point features and estimates the reliability of the final refined geometry.

\subsection{Motion Estimation and 3D Tracking}
\label{sec:method-motion}

\paragraph{Image-space tracking.}
The tracking head predicts correspondences from the geometry-conditioned motion features. We adapt CoWTracker's iterative warping head~\citep{lai2026cowtracker} for arbitrary-view tracking by removing its temporal positional embeddings. The head takes upsampled motion features concatenated with 128-channel features from a shallow convolutional RGB encoder, and retains zero-displacement initialization and visibility/confidence outputs, without assuming a chronological ordering of the input views. Outputs satisfy the query anchor $\mathbf{U}_{q\to q}(\mathbf{u})=\mathbf{u}$. Sparse query tracks are obtained by sampling the dense correspondence maps.

\paragraph{Lifting to 3D.}
Rather than directly regressing 3D motion vectors or endpoint coordinates, we obtain world-space tracks by sampling the target-view metric pointmap from Eq.~\eqref{eq:method-world-points}:
\begin{equation}
  \mathbf{Y}_{q\to i}(\mathbf{u})
  =\mathbf{X}_i^{\mathrm{metric}}\!\left(\mathbf{U}_{q\to i}(\mathbf{u})\right).
  \label{eq:method-motion-lift}
\end{equation}
This keeps correspondence regression in image space while placing every visible 3D track point on the reconstructed target-view surface.

\paragraph{Moving-object masks.}
Inspired by the foreground--background mask head of GMOS~\citep{xie2026gmos}, our query-independent mask head fuses the motion readout with features from a frozen SAM~2 image encoder~\citep{ravi2024sam2}. A convolutional decoder with higher-resolution SAM skip features predicts image-aligned moving-object probabilities. Labels describe sequence-level object motion: an object that moves within the annotated sequence remains positive in each frame where it appears, including during pauses.

\subsection{Training Objectives}
\label{sec:method-supervision}

We train the geometry branch before training the geometry-conditioned motion branch. Geometry and metric prediction are optimized jointly once metric-scale estimation is introduced. For motion training, the pretrained geometry model remains frozen, and we optimize the motion decoder and prediction heads. The geometry confidence branch is trained separately with geometry fixed. Data mixtures and the progressive training schedule are described in Section~\ref{sec:training-protocol}.

\paragraph{Geometric supervision.}
\label{sec:method-geometry-loss}
\label{sec:method-auxiliary}
We normalize both predicted and ground-truth local points by their respective sequence-wide mean valid point norms. For predictions, the factor $s(\mathbf{P})=\langle\|\mathbf{P}_i(\mathbf{u})\|_2\rangle_{i,\mathbf{u}}$ gives normalized points $\widetilde{\mathbf{P}}_i=\mathbf{P}_i/s(\mathbf{P})$. This same factor normalizes camera translations: it remains differentiable on the local-point path but is detached on the camera-translation path, as discussed in Section~\ref{sec:camera-lessons} and Eq.~\eqref{eq:camera-scale-stopgrad}. Following $\pi^3$~\citep{wang2025pi3}, we then fit an additional positive alignment scale $a$ using depth-weighted $L_1$ alignment, without a shift, using the robust, optimal, and efficient (ROE) solver of MoGe~\citep{wang2025moge}. Camera translations are multiplied by the same fitted alignment scale. Hats below denote scale-aligned predictions, with $\widehat{\mathbf{P}}_i=a\widetilde{\mathbf{P}}_i$, and stars denote normalized ground truth.

After this alignment, the local-point objective is a depth-weighted coordinate error:
\begin{equation}
  \mathcal{L}_{\mathrm{local}}
  =\left\langle
    w_i(\mathbf{u})
    \left\|\widehat{\mathbf{P}}_i(\mathbf{u})-\mathbf{P}_i^*(\mathbf{u})\right\|_1
  \right\rangle_{i,\mathbf{u}},
  \label{eq:method-local-loss}
\end{equation}
where $\langle\cdot\rangle$ averages over valid pixels and $w_i$ is a bounded inverse-depth weight. Following SurGe~\citep{knaebel2026surge}, point-gradient matching (PGM) additionally compares depth-normalized spatial point differences using a Euclidean penalty, excluding invalid neighborhoods and ground-truth depth discontinuities.

Following $\pi^3$~\citep{wang2025pi3}, $\mathcal{L}_{\mathrm{cam}}$ combines rotation-angle and $L_1$ translation errors over all distinct camera pairs. Relative poses are $\widehat{\mathbf{T}}_{j\leftarrow i}=\widehat{\mathbf{T}}_j^{-1}\widehat{\mathbf{T}}_i$. This supervision constrains inter-view geometry without fixing an absolute world frame. We also supervise points transformed into another camera's coordinates. Defining $\widehat{\mathbf{Q}}_{j\leftarrow i}=\widehat{\mathbf{T}}_{j\leftarrow i}\widehat{\mathbf{P}}_i$ and $\mathbf{Q}_{j\leftarrow i}^*=\mathbf{T}_{j\leftarrow i}^*\mathbf{P}_i^*$, the relative-point objective is
\begin{equation}
  \mathcal{L}_{\mathrm{rel}}
  =\left\langle
    w_i(\mathbf{u})
    \left\|\widehat{\mathbf{Q}}_{j\leftarrow i}(\mathbf{u})-\mathbf{Q}_{j\leftarrow i}^*(\mathbf{u})\right\|_1
  \right\rangle_{(i,j),\mathbf{u}}.
  \label{eq:method-relative-loss}
\end{equation}
This jointly constrains points and poses wherever the original pixel has valid geometry, irrespective of visibility in image $j$. On overlapping views, point-to-point consistency compares $\widehat{\mathbf{Q}}_{j\leftarrow i}$ with the predicted point sampled at its ground-truth correspondence in image $j$. Relative PGM compares spatial point differences in a common camera frame. Both use cross-view visibility masks, unlike Eq.~\eqref{eq:method-relative-loss}.

Following the released Pi3X implementation~\citep{wang2025pi3}, the ray loss $\mathcal{L}_{\mathrm{ray}}$ supervises the predicted camera-local ray coordinates $(r_i^x,r_i^y)$ in Eq.~\eqref{eq:method-local-points} against targets computed from ground-truth camera intrinsics. It is the mean absolute coordinate error over pixels with valid ray targets and does not depend on metric scale.

Inspired by the matching proxy in VGGT-$\Omega$~\citep{wang2026vggtomega}, we use a row-wise softmax objective to supervise patch matching. For a directed image pair $(i,j)$, query--key descriptor similarities are normalized with a row-wise softmax over destination patches to give $\pi_{ij}(b\mid a)$. The negative log-likelihood (NLL) is
\begin{equation}
  \mathcal{L}_{\mathrm{match}}
  =-\left\langle\log\pi_{ij}\!\left(b^*(a)\mid a\right)\right\rangle.
  \label{eq:method-match-loss}
\end{equation}
Here $a$ indexes query patches and $b^*(a)$ is the ground-truth destination patch. Only visible, in-bounds correspondences contribute.

Following the separate scale-supervision strategy of MoGe2~\citep{wang2025moge2}, we supervise the predicted metric scale against a ground-truth-derived target $m^*$ using squared error in $\log(1+m)$ space:
\begin{equation}
  \mathcal{L}_{\mathrm{metric}}
  =\left\langle
    \bigl[\log(1+m)-\log(1+m^*)\bigr]^2
  \right\rangle.
  \label{eq:method-metric-loss}
\end{equation}
The average is over sequences with native metric ground truth, and $m^*$ is treated as constant during optimization.

Let $\mathcal{L}_{\mathrm{pts}}$ denote the weighted sum of the local-point, point-gradient, relative-point, cross-view point-consistency, and relative-PGM terms. The geometry objective supervises the coarse prediction and averages point losses across refinement rounds:
\begin{equation}
  \begin{split}
  \mathcal{L}_{\mathrm{geom}}
  ={}&\mathcal{L}_{\mathrm{pts}}\!\left(\mathbf{P}^{(0)}\right)
    +\frac{1}{S}\sum_{s=1}^{S}\mathcal{L}_{\mathrm{pts}}\!\left(\mathbf{P}^{(s)}\right)\\
    &+\lambda_{\mathrm{cam}}\mathcal{L}_{\mathrm{cam}}
    +\lambda_{\mathrm{ray}}\mathcal{L}_{\mathrm{ray}}
    +\lambda_{\mathrm{match}}\mathcal{L}_{\mathrm{match}}
    +\lambda_{\mathrm{metric}}\mathcal{L}_{\mathrm{metric}}.
  \end{split}
  \label{eq:method-total-loss}
\end{equation}
Both coarse and refined predictions receive point supervision, with refined losses also propagating through the predicted rays and cameras. Refinement supervision uses reliable dense geometry; direct camera, ray, and matching terms are evaluated once on the coarse branch. Each term uses only available annotations.

\paragraph{Geometry confidence.}
Following Pi3X~\citep{wang2025pi3}, we freeze the geometry model and train the confidence branch with continuous targets that reflect reconstruction quality, rather than binary valid/invalid labels. The branch estimates per-pixel reliability of the final refined geometry.

\paragraph{Motion supervision.}
\label{sec:method-motion-loss}
We follow CoWTracker~\citep{lai2026cowtracker} for tracking supervision, combining iterative coordinate regression with binary cross-entropy for visibility and tracking confidence. We additionally supervise static-region correspondences using ground-truth geometry and cameras. Moving-object masks are supervised with binary cross-entropy on annotated moving/static pixels, ignoring unknown labels. We optimize the motion decoder and prediction heads with geometry and the SAM image encoder frozen, applying each loss only where annotations are available.

\section{Training Data and Curation}
\label{sec:training-data}

\subsection{Data Sources and Mixture}
\label{sec:data-sources}

We combine public datasets with scene-level synthetic renderings, curated Internet videos, and game-based supervision to broaden scene and camera coverage and provide complementary supervision for dynamic scenes. Motion training uses point trajectories, static geometric correspondences, and real-video masks. Table~\ref{tab:training-mixture} summarizes the data sources and sampling probabilities in the final geometry and motion training stages.

\begin{table}[!t]
\centering
\caption{\textbf{Data sources and sampling distributions.} Sampling probabilities (\%) for the geometry and motion mixtures.}\label{tab:training-mixture}
\small
\color{black}
\hypersetup{citecolor=black,linkcolor=black}
\definecolor{mixtureTrack}{HTML}{DCEAF5}
\definecolor{mixtureStatic}{HTML}{E3EEDD}
\definecolor{mixtureMask}{HTML}{F9E5D1}
\setlength{\fboxsep}{1.2pt}
\newcommand{\mixtureSplit}[3]{\tikz[baseline=(label.base)]{\node[inner xsep=1.2pt,inner ysep=1.2pt,outer sep=0pt] (label) {\phantom{#1}};\fill[#2] (label.south west) rectangle (label.north);\fill[#3] (label.south) rectangle (label.north east);\node[inner sep=0pt,anchor=base] at (label.base) {#1};}}
{\footnotesize\colorbox{mixtureTrack}{Native tracks}\qquad\colorbox{mixtureStatic}{Static reprojection tracks}\qquad\colorbox{mixtureMask}{Moving-object masks}\par}
\vspace{3pt}
\setlength{\tabcolsep}{4pt}
\setlength{\extrarowheight}{0.4pt}
\renewcommand{\arraystretch}{1.10}
\arrayrulecolor{black}
\begin{tabularx}{\linewidth}{@{}Xrr@{\hspace{14pt}}Xrr@{}}
\toprule[0.9pt]
\textbf{Dataset} & \textbf{Geometry} & \textbf{Motion} & \textbf{Dataset} & \textbf{Geometry} & \textbf{Motion} \\
\cmidrule[0.4pt](lr){1-3}\cmidrule[0.4pt](lr){4-6}
\colorbox{mixtureStatic}{ScanNet++}~\citep{yeshwanth2023scannetpp} & 3.47 & 1.06 & \colorbox{mixtureTrack}{Parallel Domain}~\citep{vanhoorick2024gcd} & 2.96 & 2.90 \\
\colorbox{mixtureStatic}{ARKitScenes}~\citep{baruch2021arkitscenes} & 1.73 & 0.53 & \colorbox{mixtureTrack}{Spring}~\citep{mehl2023spring} & 0.87 & 2.90 \\
\colorbox{mixtureStatic}{ScanNet}~\citep{dai2017scannet} & 3.04 & 0.93 & \colorbox{mixtureTrack}{Monkaa}~\citep{mayer2016sceneflow} & 0.87 & 1.45 \\
\colorbox{mixtureStatic}{BundleFusion}~\citep{dai2017bundlefusion} & 0.43 & 0.13 & \colorbox{mixtureStatic}{I2-SDF}~\citep{zhu2023i2sdf} & 0.43 & 0.13 \\
\colorbox{mixtureStatic}{RGB-D Scenes v2}~\citep{lai2014scenes} & 0.43 & 0.13 & \colorbox{mixtureTrack}{BlinkVision}~\citep{li2024blinkvision} & 0.87 & 2.90 \\
\colorbox{mixtureStatic}{12-Scenes}~\citep{valentin2016scenes} & 0.43 & 0.13 & \colorbox{mixtureStatic}{MVS-Synth}~\citep{huang2018deepmvs} & 2.96 & 0.90 \\
\colorbox{mixtureStatic}{Intrinsic3D}~\citep{maier2017intrinsic3d} & 0.43 & 0.13 & Virtual KITTI~\citep{gaidon2016virtualkitti} & 0.87 & --- \\
\colorbox{mixtureStatic}{SUN3D}~\citep{xiao2013sun3d} & 0.43 & 0.13 & \colorbox{mixtureTrack}{FlyingThings3D}~\citep{mayer2016sceneflow} & 0.87 & 2.90 \\
\colorbox{mixtureStatic}{Taskonomy}~\citep{zamir2018taskonomy} & 3.04 & 0.93 & StereoGenBench~\citep{cui2026stereogenbench} & 2.41 & --- \\
\colorbox{mixtureStatic}{Replica}~\citep{straub2019replica} & 0.43 & 0.13 & \colorbox{mixtureStatic}{MegaSynth}~\citep{jiang2025megasynth} & 0.96 & 0.29 \\
\colorbox{mixtureStatic}{ICL-NUIM}~\citep{handa2014benchmark} & 0.43 & 0.13 & \colorbox{mixtureStatic}{WHU-MVS}~\citep{liu2020whumvs} & 0.48 & 0.15 \\
\colorbox{mixtureStatic}{3RScan}~\citep{wald2019rio} & 1.45 & 0.44 & \colorbox{mixtureStatic}{TartanAir v2}~\citep{wang2020tartanair} & 0.96 & 0.29 \\
Waymo~\citep{sun2020waymo} & 2.37 & --- & \mixtureSplit{Kubric-4D}{mixtureTrack}{mixtureMask}~\citep{vanhoorick2024gcd} & 2.37 & 2.90 \\
\colorbox{mixtureStatic}{MegaDepth}~\citep{li2018megadepth} & 2.96 & 0.90 & \colorbox{mixtureTrack}{Kubric}~\citep{greff2022kubric} & 0.12 & 5.79 \\
\colorbox{mixtureStatic}{MegaDepth-X}~\citep{li2026megadepthx} & 3.04 & 0.93 & \mixtureSplit{MOVi-E}{mixtureTrack}{mixtureMask}~\citep{greff2022kubric} & 0.12 & 4.34 \\
\colorbox{mixtureStatic}{MetricScenes}~\citep{xiangli2026metricscenes} & 2.60 & 0.79 & \mixtureSplit{MOVi-F}{mixtureTrack}{mixtureMask}~\citep{greff2022kubric} & 0.12 & 4.34 \\
\colorbox{mixtureStatic}{AerialMegaDepth}~\citep{vuong2025aerialmegadepth} & 1.73 & 0.53 & \mixtureSplit{Syn4D}{mixtureTrack}{mixtureMask}~\citep{jiang2026syn4d} & 0.24 & 4.34 \\
Map-free~\citep{arnold2022mapfree} & 0.65 & --- & \colorbox{mixtureTrack}{PointOdyssey}~\citep{zheng2023pointodyssey} & 0.24 & 4.34 \\
DL3DV-10K~\citep{ling2024dl3dv} & 2.17 & --- & \colorbox{mixtureTrack}{Dynamic Replica}~\citep{karaev2023dynamicstereo} & 0.24 & 4.34 \\
\colorbox{mixtureStatic}{BlendedMVS}~\citep{yao2020blendedmvs} & 2.96 & 0.90 & \colorbox{mixtureTrack}{SynthVerse}~\citep{zhao2026synthverse} & 0.48 & 5.79 \\
\colorbox{mixtureStatic}{BlendedMVG}~\citep{yao2020blendedmvs} & 2.96 & 0.90 & Robo3R-4M~\citep{yang2026robo3r} & 0.96 & --- \\
\colorbox{mixtureStatic}{WildRGB-D}~\citep{xia2024rgbd} & 2.37 & 0.72 & \colorbox{mixtureTrack}{LSFOdyssey}~\citep{wang2024scenetracker} & --- & 0.72 \\
\colorbox{mixtureStatic}{RealX3D}~\citep{liu2025realx3d} & 0.43 & 0.13 & \mixtureSplit{GameRealm}{mixtureTrack}{mixtureStatic}~(\S\ref{sec:gamerealm-data}) & 4.50 & 20.53 \\
\colorbox{mixtureStatic}{CO3Dv2}~\citep{reizenstein2021co3d} & 2.37 & 0.72 & \colorbox{mixtureStatic}{OmniWorld}~\citep{zhou2025omniworld} & 8.28 & 0.90 \\
MVImgNet~\citep{yu2023mvimgnet} & 1.30 & --- & \colorbox{mixtureStatic}{Project Aria}~\citep{projectaria_data} & 3.47 & 1.06 \\
EPIC-KITCHENS~\citep{damen2018epic} & 2.96 & --- & Aria Digital Twin~\citep{pan2023adt} & 1.00 & --- \\
\colorbox{mixtureStatic}{TartanAir}~\citep{wang2020tartanair} & 2.96 & 0.90 & \colorbox{mixtureStatic}{Funes}~(\S\ref{sec:synthetic-data}) & 0.95 & 0.29 \\
\colorbox{mixtureStatic}{GTA-SfM}~\citep{wang2020flowmotion} & 2.96 & 0.90 & \colorbox{mixtureStatic}{Sketchfab}~(\S\ref{sec:synthetic-data}) & 6.57 & 2.00 \\
\colorbox{mixtureStatic}{MatrixCity}~\citep{li2023matrixcity} & 0.43 & 0.13 & Internet videos~(\S\ref{sec:internet-video-curation}) & 1.00 & --- \\
\colorbox{mixtureStatic}{Mid-Air}~\citep{fonder2019midair} & 0.87 & 0.26 & \colorbox{mixtureMask}{GMOS-2K}~\citep{xie2026gmos} & --- & 10.00 \\
\bottomrule[0.9pt]
\end{tabularx}
\arrayrulecolor{black}
\par\vspace{4pt}
\begin{minipage}{\linewidth}
\footnotesize\raggedright
\textbf{Labels.} Blue: native tracks; green: static RGB-D reprojection tracks; orange: explicit moving-object masks. Split backgrounds indicate multiple types. Labels may cover only part of a grouped dataset; all-static negatives are not marked as moving-object masks.\par
\textbf{Sampling.} Probabilities, not frame counts; each stage sums to 100\%. Motion: 70.2\% tracking, 19.8\% static geometry, and 10.0\% moving-object masks from real videos.
\end{minipage}
\end{table}

\subsection{Scene-Level Synthetic Data}
\label{sec:synthetic-data}

\paragraph{Scene assets.}
We curate approximately 10,000 scene-level 3D assets from Funes~\citep{funesworld} and Sketchfab~\citep{sketchfab} for synthetic multi-view rendering. These assets include architectural environments and other large-scale scenes, rather than only isolated objects. They support exterior observations and, where the geometry permits, interior views. Our goal is to broaden the diversity of scene layouts and camera configurations represented during training, rather than simply increase the number of rendered images.

\paragraph{Candidate views and validation.}
We generate candidate cameras across positions, elevations, viewing directions, and distances using surrounding-space and surface-guided proposals. Exterior views capture both overall structure and local detail; interior views require sufficient free space and renderable surfaces. Visibility and RGB-D quality checks reject invalid cameras and degenerate observations, including views dominated by open mesh backsides.

\paragraph{Coverage-aware views and connected chains.}
We select views by their incremental coverage of observable surfaces, balancing surface area with spatial and orientation diversity so that large planar regions do not dominate. Depth-consistent reprojections then define a view-overlap graph, from which we sample connected view chains. Neighboring views retain reliable common geometry, while distant views may have little or no direct overlap. This design decouples global scene coverage from local cross-view connectivity: views need not all overlap a single reference or follow a smooth temporal trajectory.

\begin{figure}[!htbp]
  \centering
  \includegraphics[width=\linewidth]{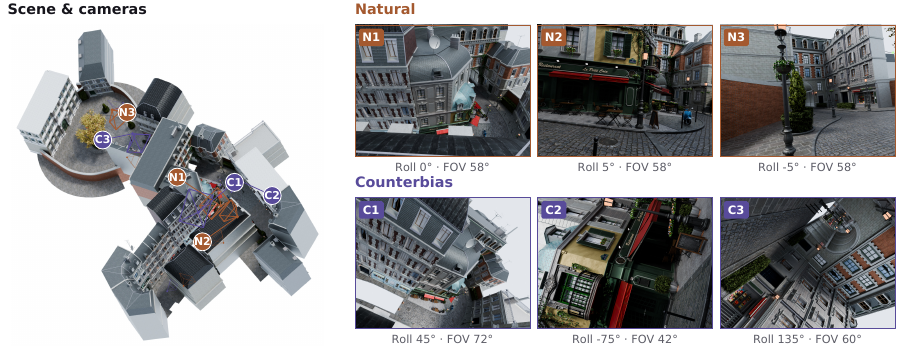}
  \caption{\textbf{Scene-level view sampling.} Labeled camera frusta correspond to the rendered views. Natural views use small roll and shared intrinsics; independently positioned counterbias views span broader roll and variable focal length. Labels report roll and vertical FOV. Illustrative camera configurations. Scene: Amazon Lumberyard Bistro~\citep{ORCAAmazonBistro}, \href{https://creativecommons.org/licenses/by/4.0/}{CC BY 4.0}; custom cameras, lighting, and overlays.}
  \label{fig:synthetic-view-sampling}
\end{figure}

\paragraph{Camera-distribution coverage.}
We construct independent \emph{natural} and \emph{counterbias} view sets. Natural sets use modest camera roll and shared intrinsics, whereas counterbias sets explicitly cover broader roll angles and vary focal length across views. The two sets use independently sampled camera positions, diversifying viewpoints and occlusions rather than forming view-by-view pairs. Both sets follow the same geometric-validity principles. RGB and depth are rendered with the assigned camera parameters, followed by final quality validation and depth-based overlap computation. Figure~\ref{fig:synthetic-view-sampling} illustrates the two camera-sampling regimes on a scene-level asset.

\FloatBarrier

\subsection{Sequence-Consistent Projection Augmentation}
\label{sec:projection-augmentation}

Geometry training data are dominated by pinhole or rectified imagery, limiting the camera projections encountered during training. To counter this bias, we apply sequence-consistent lens and crop-and-resize augmentation. Each sequence receives an identity, barrel, blended equidistant (fisheye-like), or pincushion mapping with probability 40\%, 35\%, 20\%, or 5\%, respectively. Lens parameters and the final crop and resize are shared across frames.

The augmentation transforms RGB and prepared geometric supervision together. For each output pixel, an inverse warp identifies a source location whose calibrated ray and camera-axis depth define the new point target, without approximating the distorted view with a new pinhole camera. Correspondence targets, visibility, and cross-view overlap are recomputed under the same mapping, while camera poses and scene scale remain unchanged. We crop to valid source-image support and resize, reducing distortion strength when necessary to satisfy a zoom limit.

These transforms diversify image-to-ray mappings within the observed field of view, complementing genuine camera diversity.

\Needspace{16\baselineskip}
\subsection{Foreground-Suppressed Supervision from Internet Videos}
\label{sec:internet-video-curation}

\begingroup
\setlength{\columnsep}{6pt}
\setlength{\intextsep}{2pt}
\begin{wrapfigure}{r}{0.49\textwidth}
  \centering
  \setlength{\abovecaptionskip}{2pt}
  \includegraphics[width=\linewidth]{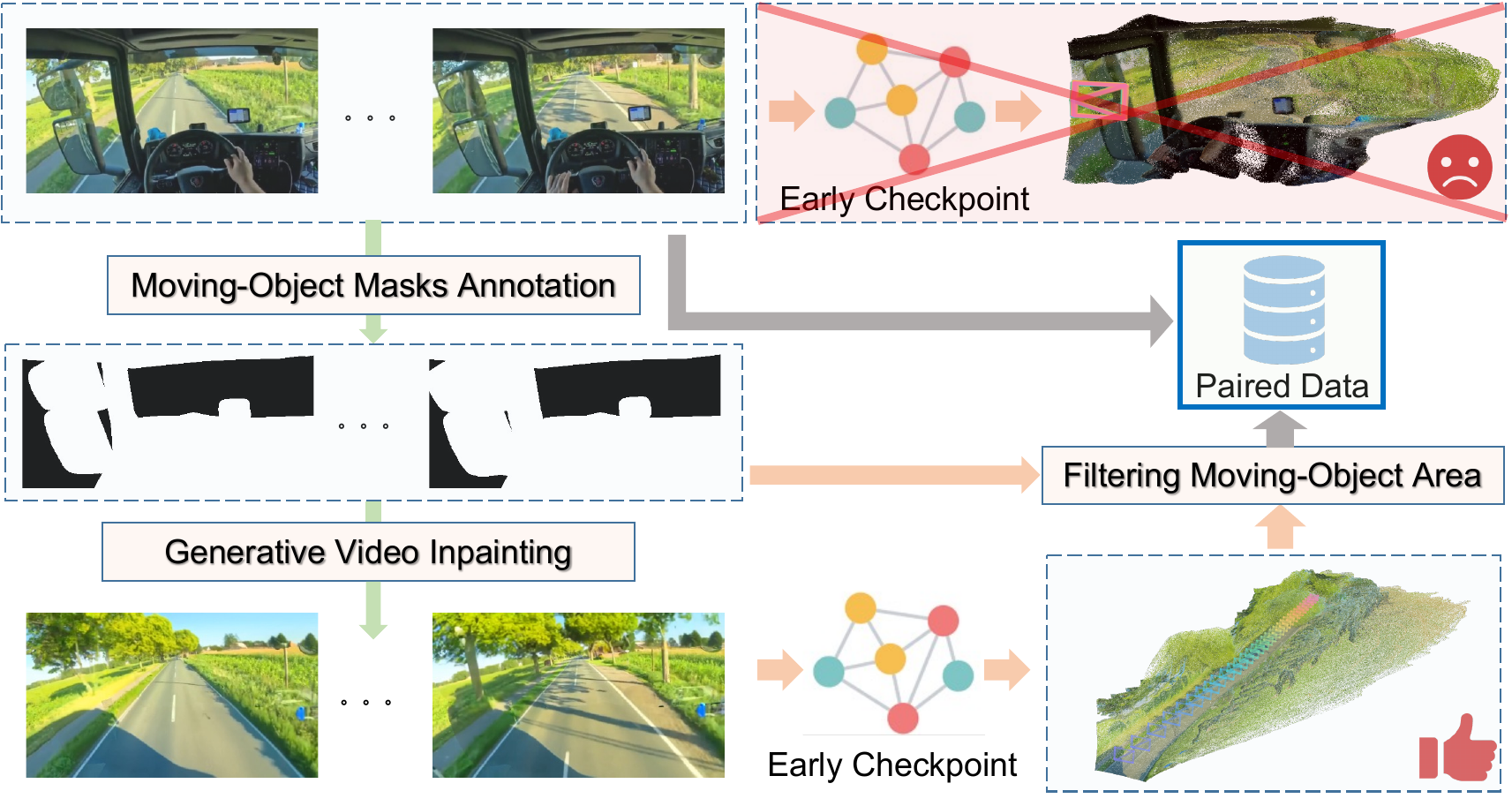}
  \caption{\textbf{Internet-video data pipeline.} Background geometry is reconstructed and filtered with moving-object masks.}
  \label{fig:foreground-pseudo-labeling}
  \vspace{-\baselineskip}
\end{wrapfigure}

\paragraph{Foreground-dominated reconstruction failures.}
Large moving foregrounds in Internet videos can disrupt geometry reconstruction, especially in egocentric sequences where the foreground dominates the view. These failures motivate suppressing the foreground and completing the video before geometry inference.

\paragraph{Foreground-suppressed pseudo-labeling.}
We first mask moving foreground regions and complete the video using generative video inpainting~\citep{wan2025video,alibabapaiwan21fun}. An earlier geometry-only checkpoint of \nickname{} then infers a background reconstruction from the completed video. We filter the resulting point clouds using the moving-object masks and retain only points corresponding to originally visible background pixels. This avoids placing pseudo-labels directly on inpainted content. We pair these pseudo-labels with the original RGB frames, rather than the completed video, for subsequent geometry training. Figure~\ref{fig:foreground-pseudo-labeling} summarizes this pipeline.
\par
\endgroup

\subsection{GameRealm: Game-Based Geometry and Motion}
\label{sec:gamerealm-data}

To broaden supervision for joint geometry and motion learning in dynamic environments, we use \emph{GameRealm}~\citep{gamerealm}, a game-based dataset developed jointly with collaborators. It provides engine-native geometric and motion annotations, including posed multi-view RGB-D observations for reconstruction and persistent 2D/3D point trajectories for tracking. Covering diverse scenes, viewpoints, characters, and interactions, these data complement our static scene renderings and real-video pseudo-labels with supervision for both spatial reconstruction and temporal correspondence.

\subsection{Training Protocol}
\label{sec:training-protocol}

\nickname{} is trained progressively on 64 NVIDIA A800 GPUs, with approximately 300,000 optimizer updates across geometry-training stages. Starting from a pretrained DINOv3 encoder, early stages train the image encoder, looped decoder, and point, camera, and matching heads. Later stages introduce sparse 3D refinement and metric-scale estimation while expanding the data mixture with curated Internet videos. Successive stages inherit the preceding model weights.

Separately, motion training spans approximately 100,000 updates, including joint optimization of the motion decoder, tracking head, and moving-object mask head, with geometry frozen throughout. The SAM image encoder remains frozen during mask training, and the confidence head is trained separately with geometry fixed.

Optimization uses AdamW and BF16 mixed precision. In the final geometry stage, images have a long side of 512 pixels and varying aspect ratios, with at most 64 images per GPU and no gradient accumulation. For sets of $K$ views (Section~\ref{sec:training-view-counts}), each GPU packs $\lfloor 64/K \rfloor$ independent examples per batch. The base and encoder learning rates are $1.5\times10^{-5}$ and $2\times10^{-6}$, respectively, with cosine decay. All update counts denote optimizer steps under weighted data sampling, rather than dataset epochs.

\FloatBarrier
\section{Evaluation}
\label{sec:evaluation}

\subsection{Geometry Evaluation}
\label{sec:geometry-evaluation}

\paragraph{Standard benchmarks.}
Table~\ref{tab:geometry-results} compares \nickname{} with nine baselines across 16 settings spanning depth~\citep{butler2012sintel,palazzolo2019refusion,geiger2012kitti,silberman2012nyuv2}, pose and trajectory~\citep{zhou2018stereo,reizenstein2021co3d,sturm2012tum,dai2017scannet}, and multiview reconstruction~\citep{shotton2013scenes,azinovic2022neuralrgbd,jensen2014dtu,schops2017eth3d}. Monocular depth uses per-frame median-scale alignment, whereas video depth uses sequence-level scale-and-shift alignment. The full geometry-and-motion system has approximately 0.5B parameters, including frozen modules; Table~\ref{tab:geometry-results} instead reports 343M loaded parameters for the geometry model used in this comparison.

\nickname{} achieves the best overall average rank (3.00, compared with 3.10 for Pi3X), with the best panel-average ranks for video depth, pose/trajectory, and multiview reconstruction. It attains the lowest Sintel video-depth AbsRel and the lowest reconstruction accuracy error on DTU and ETH3D. Metric-scale performance is evaluated separately in Table~\ref{tab:geometry-results}.

\begin{figure}[!htb]
  \centering
  \includegraphics[width=\linewidth]{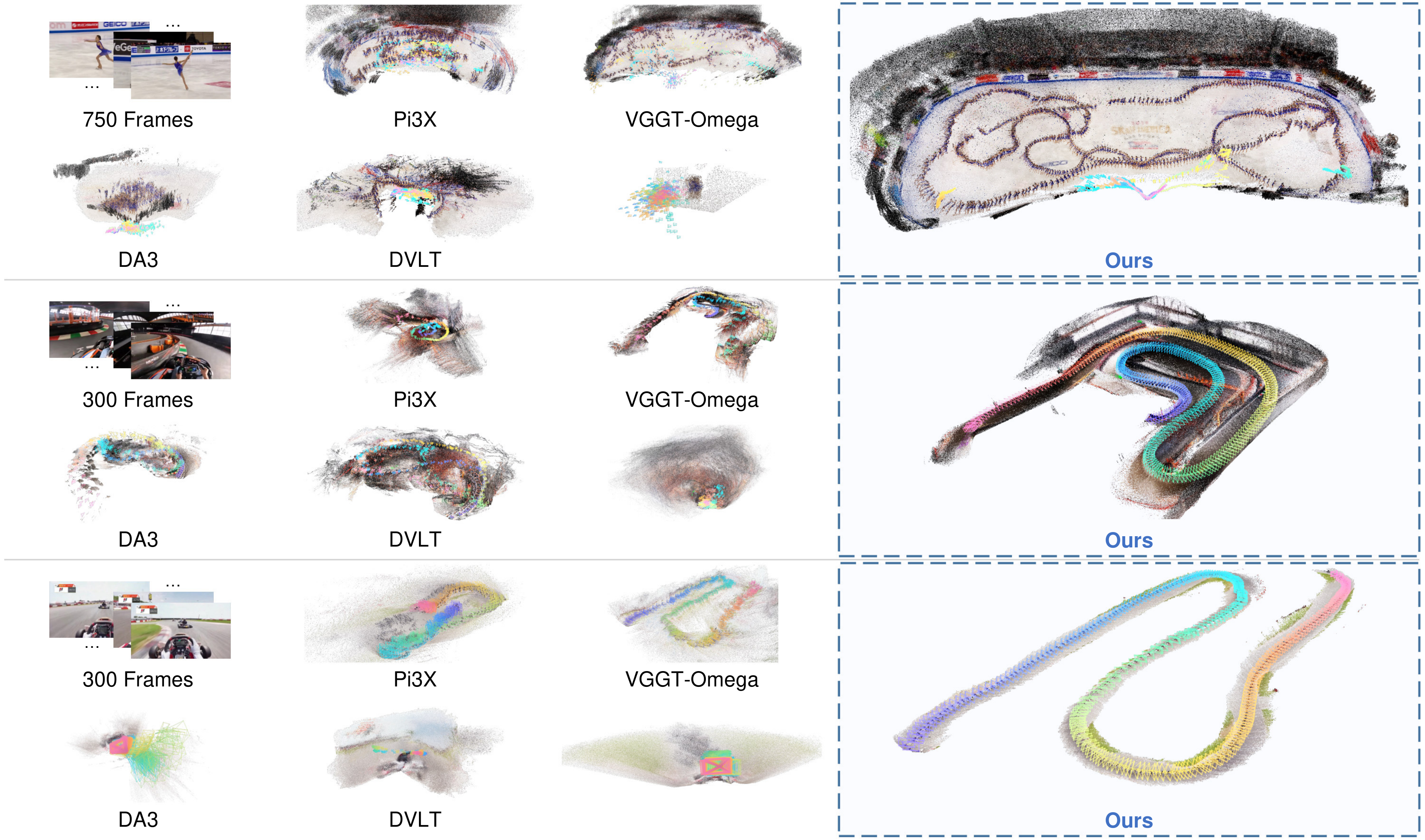}
  \caption{\textbf{Qualitative reconstruction comparison.} Reconstructed point clouds and camera poses.}
  \label{fig:reconstruction-comparison}
\end{figure}

\begin{table}[!p]
\centering
\begingroup
\definecolor{geomaccent}{HTML}{174B68}
\definecolor{geomdivider}{HTML}{B7BDC3}
\definecolor{geomsecondary}{HTML}{65717C}
\definecolor{tabfirst}{rgb}{1,0.7,0.7}
\definecolor{tabsecond}{rgb}{1,0.85,0.7}
\definecolor{tabthird}{rgb}{1,1,0.7}
\colorlet{tabfirst}{tabfirst!55!white}
\colorlet{tabsecond}{tabsecond!55!white}
\colorlet{tabthird}{tabthird!55!white}
\setlength{\tabcolsep}{0.35pt}
\renewcommand{\arraystretch}{1}
\setlength{\parskip}{0pt}
\setlength{\aboverulesep}{0.7pt}
\setlength{\belowrulesep}{0.7pt}
\setlength{\cmidrulesep}{0pt}
\setlength{\cmidrulekern}{1pt}
\fontsize{7}{8.5}\selectfont
\noindent\textcolor{geomaccent}{\textbf{4 EVALUATION GROUPS\quad 16 SETTINGS\quad 10 METHODS}}\hfill\textbf{1st overall $\cdot$ 3/4 group leads}
\par\vspace{1.5pt}
\fontsize{7.2}{8.4}\selectfont
\begin{tabularx}{\linewidth}[t]{@{}l*{9}{>{\centering\arraybackslash}X}!{\hspace{2pt}{\color{geomdivider}\vrule width 0.3pt}\hspace{2pt}}>{\centering\arraybackslash}X@{}}
\toprule
\multicolumn{11}{@{}l}{\color{geomaccent}\bfseries Average ranks $\downarrow$} \\
Evaluation & {\fontsize{5.9}{7.2}\selectfont VGGT} & {\fontsize{5.9}{7.2}\selectfont $\pi^3$} & {\fontsize{5.9}{7.2}\selectfont MapAnything} & {\fontsize{5.9}{7.2}\selectfont DA3-Giant 1.1} & {\fontsize{5.9}{7.2}\selectfont Pi3X} & {\fontsize{5.9}{7.2}\selectfont VGGT-$\Omega$} & {\fontsize{5.9}{7.2}\selectfont DVLT} & {\fontsize{5.9}{7.2}\selectfont ARROW} & {\fontsize{5.9}{7.2}\selectfont Track4World} & {\fontsize{5.9}{7.2}\selectfont \nickname{}} \\[-1.2pt]
{\fontsize{5.8}{7.2}\selectfont\textcolor{geomsecondary}{Params.}} & {\fontsize{5.8}{7.2}\selectfont\textcolor{geomsecondary}{1.26B}} & {\fontsize{5.8}{7.2}\selectfont\textcolor{geomsecondary}{959M}} & {\fontsize{5.8}{7.2}\selectfont\textcolor{geomsecondary}{1.23B}} & {\fontsize{5.8}{7.2}\selectfont\textcolor{geomsecondary}{1.36B}} & {\fontsize{5.8}{7.2}\selectfont\textcolor{geomsecondary}{992M}} & {\fontsize{5.8}{7.2}\selectfont\textcolor{geomsecondary}{1.14B}} & {\fontsize{5.8}{7.2}\selectfont\textcolor{geomsecondary}{117M}} & {\fontsize{5.8}{7.2}\selectfont\textcolor{geomsecondary}{1.32B}} & {\fontsize{5.8}{7.2}\selectfont\textcolor{geomsecondary}{1.72B}} & {\fontsize{5.8}{7.2}\selectfont\textcolor{geomsecondary}{343M}} \\
\midrule
\hyperlink{geometry-monodepth}{Mono. (a)} & 7.39 & \cellcolor{tabsecond}2.96 & 6.14 & 6.89 & \cellcolor{tabfirst}2.57 & 6.75 & 5.57 & 6.68 & 6.54 & \cellcolor{tabthird}3.50 \\
\hyperlink{geometry-videodepth}{Video (b)} & 8.62 & 4.10 & 8.29 & 6.67 & \cellcolor{tabsecond}3.10 & 6.90 & 6.86 & 4.29 & \cellcolor{tabthird}3.86 & \cellcolor{tabfirst}2.33 \\
\hyperlink{geometry-pose-trajectory}{Pose \& Traj. (c)} & 8.05 & 4.50 & 9.50 & \cellcolor{tabsecond}3.31 & \cellcolor{tabsecond}3.31 & \cellcolor{tabthird}3.83 & 5.81 & 6.71 & 6.98 & \cellcolor{tabfirst}3.00 \\
\hyperlink{geometry-mv-recon}{Recon. (d)} & 7.00 & \cellcolor{tabthird}3.83 & 7.92 & 5.25 & \cellcolor{tabsecond}3.42 & 6.17 & 5.50 & 5.08 & 7.67 & \cellcolor{tabfirst}3.17 \\
\midrule
\textbf{Overall} & 7.76 & \cellcolor{tabthird}3.85 & 7.96 & 5.53 & \cellcolor{tabsecond}3.10 & 5.91 & 5.93 & 5.69 & 6.26 & \cellcolor{tabfirst}3.00 \\
\textbf{Rank} & 9 & \cellcolor{tabthird}3 & 10 & 4 & \cellcolor{tabsecond}2 & 6 & 7 & 5 & 8 & \cellcolor{tabfirst}1 \\
\bottomrule
\end{tabularx}
\par\vspace{2pt}
\fontsize{5.9}{7.2}\selectfont
\begin{tabular*}{\linewidth}[t]{@{\extracolsep{\fill}}l*{28}{c}@{}}
\toprule
\multicolumn{29}{@{}l}{\color{geomaccent}\bfseries \hypertarget{geometry-monodepth}{(a) Monocular depth}} \\
 & \multicolumn{7}{c}{\textbf{Sintel}} & \multicolumn{7}{c}{\textbf{Bonn}} & \multicolumn{7}{c}{\textbf{KITTI}} & \multicolumn{7}{c}{\textbf{NYU-v2}} \\
\cmidrule(lr){2-8}\cmidrule(lr){9-15}\cmidrule(lr){16-22}\cmidrule(lr){23-29}
Method & Rel$\downarrow$ & SqRel$\downarrow$ & RMS$\downarrow$ & log$\downarrow$ & $\delta_1\!\uparrow$ & $\delta_2\!\uparrow$ & $\delta_3\!\uparrow$ & Rel$\downarrow$ & SqRel$\downarrow$ & RMS$\downarrow$ & log$\downarrow$ & $\delta_1\!\uparrow$ & $\delta_2\!\uparrow$ & $\delta_3\!\uparrow$ & Rel$\downarrow$ & SqRel$\downarrow$ & RMS$\downarrow$ & log$\downarrow$ & $\delta_1\!\uparrow$ & $\delta_2\!\uparrow$ & $\delta_3\!\uparrow$ & Rel$\downarrow$ & SqRel$\downarrow$ & RMS$\downarrow$ & log$\downarrow$ & $\delta_1\!\uparrow$ & $\delta_2\!\uparrow$ & $\delta_3\!\uparrow$ \\
\midrule
VGGT & .335 & 2.131 & 5.434 & .449 & 59.8 & 75.1 & 84.7 & .053 & .040 & .255 & .112 & 97.0 & 98.1 & 98.8 & .082 & .402 & 3.651 & .124 & 94.7 & 98.4 & 99.3 & \cellcolor{tabthird}.056 & .045 & .308 & .110 & \cellcolor{tabthird}95.1 & 98.1 & \cellcolor{tabthird}99.1 \\
$\pi^3$ & \cellcolor{tabsecond}.280 & \cellcolor{tabsecond}1.280 & \cellcolor{tabsecond}3.716 & .528 & 61.6 & 78.7 & 85.1 & \cellcolor{tabthird}.048 & \cellcolor{tabsecond}.027 & \cellcolor{tabthird}.206 & \cellcolor{tabthird}.095 & \cellcolor{tabsecond}97.4 & \cellcolor{tabsecond}98.7 & \cellcolor{tabfirst}99.2 & \cellcolor{tabsecond}.059 & .313 & 3.127 & \cellcolor{tabfirst}.097 & \cellcolor{tabfirst}97.1 & \cellcolor{tabfirst}99.2 & \cellcolor{tabfirst}99.7 & \cellcolor{tabsecond}.054 & \cellcolor{tabsecond}.036 & \cellcolor{tabsecond}.290 & \cellcolor{tabsecond}.103 & \cellcolor{tabsecond}95.6 & \cellcolor{tabsecond}98.6 & \cellcolor{tabsecond}99.4 \\
MapAnything & .291 & 1.931 & 4.356 & .428 & \cellcolor{tabsecond}64.7 & \cellcolor{tabthird}80.3 & \cellcolor{tabthird}87.2 & .068 & .046 & .238 & .116 & 96.4 & 98.3 & \cellcolor{tabthird}99.0 & \cellcolor{tabthird}.062 & .333 & 3.259 & .106 & \cellcolor{tabthird}96.3 & 98.8 & \cellcolor{tabsecond}99.6 & .067 & .052 & .329 & .116 & 94.1 & 97.9 & \cellcolor{tabthird}99.1 \\
DA3-Giant 1.1 & .359 & 2.364 & 4.745 & .447 & 60.6 & 76.6 & 84.3 & .054 & .040 & .225 & .104 & 96.9 & 98.3 & \cellcolor{tabthird}99.0 & .077 & \cellcolor{tabthird}.308 & \cellcolor{tabsecond}3.062 & .113 & 95.8 & \cellcolor{tabthird}99.0 & \cellcolor{tabfirst}99.7 & .087 & .101 & .533 & .281 & 90.3 & 94.2 & 95.8 \\
Pi3X & \cellcolor{tabthird}.284 & \cellcolor{tabthird}1.389 & \cellcolor{tabthird}3.842 & .498 & 61.6 & 77.3 & 86.0 & \cellcolor{tabfirst}.041 & \cellcolor{tabfirst}.026 & \cellcolor{tabfirst}.198 & \cellcolor{tabfirst}.087 & \cellcolor{tabfirst}97.7 & \cellcolor{tabfirst}98.8 & \cellcolor{tabfirst}99.2 & \cellcolor{tabfirst}.053 & .309 & 3.222 & \cellcolor{tabsecond}.098 & \cellcolor{tabfirst}97.1 & \cellcolor{tabsecond}99.1 & \cellcolor{tabsecond}99.6 & \cellcolor{tabfirst}.052 & \cellcolor{tabfirst}.031 & \cellcolor{tabfirst}.269 & \cellcolor{tabfirst}.096 & \cellcolor{tabfirst}95.8 & \cellcolor{tabfirst}98.7 & \cellcolor{tabfirst}99.5 \\
VGGT-$\Omega$ & .307 & 1.605 & 4.024 & \cellcolor{tabthird}.401 & 62.7 & 78.5 & 86.8 & .117 & .143 & .319 & .165 & 93.0 & 95.3 & 96.6 & .103 & .442 & 3.940 & .139 & 92.8 & 98.6 & \cellcolor{tabthird}99.5 & .058 & \cellcolor{tabthird}.041 & \cellcolor{tabthird}.299 & \cellcolor{tabthird}.104 & 95.0 & \cellcolor{tabthird}98.4 & \cellcolor{tabsecond}99.4 \\
DVLT & .319 & 1.860 & 4.401 & \cellcolor{tabsecond}.393 & 62.1 & \cellcolor{tabsecond}81.5 & \cellcolor{tabfirst}88.8 & .060 & .040 & .230 & .107 & 96.6 & 98.4 & \cellcolor{tabthird}99.0 & .079 & .402 & 3.330 & .120 & 95.5 & 98.8 & 99.4 & .064 & .043 & .301 & \cellcolor{tabthird}.104 & 94.8 & \cellcolor{tabsecond}98.6 & \cellcolor{tabfirst}99.5 \\
ARROW & .365 & 2.032 & 4.181 & .637 & 55.7 & 72.8 & 80.1 & .059 & .037 & .224 & .105 & 97.1 & 98.5 & \cellcolor{tabsecond}99.1 & .065 & \cellcolor{tabsecond}.306 & \cellcolor{tabfirst}2.917 & .105 & 96.2 & 98.9 & \cellcolor{tabsecond}99.6 & .087 & .112 & .544 & .309 & 90.7 & 94.9 & 96.4 \\
Track4World & .329 & 1.935 & 4.363 & .412 & \cellcolor{tabthird}63.2 & 78.3 & 86.2 & \cellcolor{tabsecond}.047 & \cellcolor{tabthird}.030 & \cellcolor{tabsecond}.203 & \cellcolor{tabsecond}.092 & \cellcolor{tabthird}97.3 & \cellcolor{tabthird}98.6 & \cellcolor{tabfirst}99.2 & .079 & .358 & 3.514 & .126 & 95.0 & 98.6 & 99.4 & .089 & .121 & .570 & .372 & 90.1 & 93.9 & 95.4 \\
\midrule
\nickname{} & \cellcolor{tabfirst}.256 & \cellcolor{tabfirst}1.011 & \cellcolor{tabfirst}3.192 & \cellcolor{tabfirst}.386 & \cellcolor{tabfirst}68.0 & \cellcolor{tabfirst}81.7 & \cellcolor{tabsecond}88.1 & .055 & .035 & .223 & .101 & \cellcolor{tabthird}97.3 & \cellcolor{tabthird}98.6 & \cellcolor{tabsecond}99.1 & \cellcolor{tabthird}.062 & \cellcolor{tabfirst}.286 & \cellcolor{tabthird}3.079 & \cellcolor{tabthird}.100 & \cellcolor{tabsecond}96.9 & \cellcolor{tabfirst}99.2 & \cellcolor{tabfirst}99.7 & .064 & .047 & .328 & .117 & 94.1 & 97.9 & \cellcolor{tabthird}99.1 \\
\bottomrule
\end{tabular*}
\par\vspace{1.5pt}
\fontsize{6.25}{7.2}\selectfont
\begin{tabular*}{\linewidth}[t]{@{\extracolsep{\fill}}l*{21}{c}@{}}
\toprule
\multicolumn{22}{@{}l}{\color{geomaccent}\bfseries \hypertarget{geometry-videodepth}{(b) Video depth}} \\
 & \multicolumn{7}{c}{\textbf{Sintel}} & \multicolumn{7}{c}{\textbf{Bonn}} & \multicolumn{7}{c}{\textbf{KITTI}} \\
\cmidrule(lr){2-8}\cmidrule(lr){9-15}\cmidrule(lr){16-22}
Method & Rel$\downarrow$ & SqRel$\downarrow$ & RMS$\downarrow$ & log$\downarrow$ & $\delta_1\!\uparrow$ & $\delta_2\!\uparrow$ & $\delta_3\!\uparrow$ & Rel$\downarrow$ & SqRel$\downarrow$ & RMS$\downarrow$ & log$\downarrow$ & $\delta_1\!\uparrow$ & $\delta_2\!\uparrow$ & $\delta_3\!\uparrow$ & Rel$\downarrow$ & SqRel$\downarrow$ & RMS$\downarrow$ & log$\downarrow$ & $\delta_1\!\uparrow$ & $\delta_2\!\uparrow$ & $\delta_3\!\uparrow$ \\
\midrule
VGGT & .230 & 1.743 & 5.654 & .407 & 67.8 & 81.3 & 89.7 & .052 & .036 & .276 & .123 & 96.9 & 98.0 & 98.7 & .052 & .311 & 3.299 & .104 & 96.8 & 98.7 & 99.4 \\
$\pi^3$ & .211 & 1.290 & \cellcolor{tabsecond}4.002 & .484 & \cellcolor{tabthird}72.7 & 84.0 & 87.4 & .044 & .028 & .224 & .102 & 97.5 & 98.6 & \cellcolor{tabthird}99.1 & \cellcolor{tabthird}.037 & \cellcolor{tabsecond}.174 & \cellcolor{tabthird}2.491 & \cellcolor{tabfirst}.074 & \cellcolor{tabsecond}98.5 & \cellcolor{tabfirst}99.5 & \cellcolor{tabfirst}99.8 \\
MapAnything & .240 & 1.482 & 4.916 & .454 & 65.1 & 79.7 & 86.3 & .050 & .042 & .241 & .110 & 97.5 & 98.4 & 99.0 & .056 & .275 & 3.034 & .099 & 97.3 & 99.0 & \cellcolor{tabthird}99.6 \\
DA3-Giant 1.1 & .210 & 1.587 & 4.994 & .389 & 69.8 & 81.9 & 87.1 & .048 & .036 & .228 & .104 & 97.3 & 98.5 & 99.0 & .053 & .206 & 2.707 & .090 & 97.6 & \cellcolor{tabthird}99.3 & \cellcolor{tabfirst}99.8 \\
Pi3X & \cellcolor{tabthird}.189 & \cellcolor{tabsecond}1.215 & \cellcolor{tabthird}4.043 & .460 & \cellcolor{tabsecond}73.7 & 83.8 & 87.3 & \cellcolor{tabfirst}.038 & \cellcolor{tabsecond}.026 & \cellcolor{tabthird}.212 & \cellcolor{tabthird}.095 & 97.5 & \cellcolor{tabthird}98.7 & \cellcolor{tabsecond}99.2 & \cellcolor{tabfirst}.034 & \cellcolor{tabfirst}.140 & \cellcolor{tabsecond}2.310 & \cellcolor{tabsecond}.076 & \cellcolor{tabfirst}98.6 & \cellcolor{tabfirst}99.5 & \cellcolor{tabfirst}99.8 \\
VGGT-$\Omega$ & .201 & \cellcolor{tabthird}1.257 & 4.566 & .384 & 63.9 & 81.3 & \cellcolor{tabthird}92.3 & .117 & .157 & .353 & .185 & 93.2 & 94.9 & 96.1 & .049 & .201 & 2.727 & .088 & 98.0 & \cellcolor{tabthird}99.3 & \cellcolor{tabsecond}99.7 \\
DVLT & .211 & 1.448 & 4.889 & \cellcolor{tabsecond}.359 & 70.0 & 84.9 & 90.9 & .045 & .034 & .229 & .103 & 97.4 & 98.6 & \cellcolor{tabthird}99.1 & .060 & .338 & 3.173 & .108 & 97.3 & 98.8 & 99.4 \\
ARROW & .238 & 2.924 & 4.718 & .416 & 69.7 & \cellcolor{tabthird}86.2 & \cellcolor{tabfirst}93.3 & \cellcolor{tabthird}.042 & .029 & .216 & .096 & \cellcolor{tabthird}97.8 & \cellcolor{tabthird}98.7 & \cellcolor{tabsecond}99.2 & \cellcolor{tabfirst}.034 & .185 & \cellcolor{tabfirst}2.195 & \cellcolor{tabsecond}.076 & \cellcolor{tabthird}98.4 & \cellcolor{tabsecond}99.4 & \cellcolor{tabsecond}99.7 \\
Track4World & \cellcolor{tabsecond}.181 & 1.369 & 4.106 & \cellcolor{tabthird}.367 & 71.9 & \cellcolor{tabfirst}86.5 & \cellcolor{tabsecond}92.9 & \cellcolor{tabsecond}.039 & \cellcolor{tabthird}.027 & \cellcolor{tabfirst}.201 & \cellcolor{tabfirst}.089 & \cellcolor{tabsecond}98.0 & \cellcolor{tabsecond}98.8 & \cellcolor{tabfirst}99.3 & .039 & .206 & 2.792 & .091 & 97.9 & 99.2 & \cellcolor{tabthird}99.6 \\
\midrule
\nickname{} & \cellcolor{tabfirst}.160 & \cellcolor{tabfirst}.653 & \cellcolor{tabfirst}3.142 & \cellcolor{tabfirst}.344 & \cellcolor{tabfirst}77.0 & \cellcolor{tabsecond}86.3 & 90.2 & .043 & \cellcolor{tabfirst}.024 & \cellcolor{tabsecond}.205 & \cellcolor{tabsecond}.091 & \cellcolor{tabfirst}98.1 & \cellcolor{tabfirst}98.9 & \cellcolor{tabfirst}99.3 & \cellcolor{tabsecond}.036 & \cellcolor{tabthird}.177 & 2.571 & \cellcolor{tabthird}.077 & \cellcolor{tabthird}98.4 & \cellcolor{tabsecond}99.4 & \cellcolor{tabsecond}99.7 \\
\bottomrule
\end{tabular*}
\par\vspace{1.5pt}
\fontsize{6.25}{7.2}\selectfont
\begin{tabular*}{\linewidth}[t]{@{\extracolsep{\fill}}l*{21}{c}@{}}
\toprule
\multicolumn{22}{@{}l}{\color{geomaccent}\bfseries \hypertarget{geometry-pose-trajectory}{(c) Relative pose and camera trajectory}} \\
 & \multicolumn{6}{c}{\textbf{RealEstate10K}} & \multicolumn{6}{c}{\textbf{CO3Dv2}} & \multicolumn{3}{c}{\textbf{Sintel}} & \multicolumn{3}{c}{\textbf{TUM}} & \multicolumn{3}{c}{\textbf{ScanNetv2}} \\
\cmidrule(lr){2-7}\cmidrule(lr){8-13}\cmidrule(lr){14-16}\cmidrule(lr){17-19}\cmidrule(lr){20-22}
Method & $R_{5}\!\uparrow$ & $T_{5}\!\uparrow$ & $A_{5}\!\uparrow$ & $R_{30}\!\uparrow$ & $T_{30}\!\uparrow$ & $A_{30}\!\uparrow$ & $R_{5}\!\uparrow$ & $T_{5}\!\uparrow$ & $A_{5}\!\uparrow$ & $R_{30}\!\uparrow$ & $T_{30}\!\uparrow$ & $A_{30}\!\uparrow$ & ATE$\downarrow$ & RPE$_t\!\downarrow$ & RPE$_r\!\downarrow$ & ATE$\downarrow$ & RPE$_t\!\downarrow$ & RPE$_r\!\downarrow$ & ATE$\downarrow$ & RPE$_t\!\downarrow$ & RPE$_r\!\downarrow$ \\
\midrule
VGGT & 95.1 & 53.9 & 30.4 & \cellcolor{tabfirst}100.0 & 92.0 & 74.5 & 95.0 & 85.1 & 60.3 & 99.0 & 97.1 & 88.6 & .1671 & .0622 & .4906 & .0119 & .0102 & .3112 & .0352 & .0153 & .3808 \\
$\pi^3$ & \cellcolor{tabfirst}98.8 & 79.6 & 62.7 & \cellcolor{tabfirst}100.0 & 95.6 & 87.2 & \cellcolor{tabsecond}97.3 & 86.8 & \cellcolor{tabthird}67.1 & 99.1 & \cellcolor{tabthird}97.4 & 90.2 & \cellcolor{tabthird}.0733 & .0390 & \cellcolor{tabthird}.2773 & .0142 & .0090 & .3114 & \cellcolor{tabthird}.0305 & .0126 & \cellcolor{tabsecond}.3464 \\
MapAnything & 98.4 & 65.7 & 45.0 & \cellcolor{tabfirst}100.0 & 93.2 & 79.9 & 64.1 & 59.0 & 24.7 & 96.1 & 92.3 & 71.6 & .2182 & .0791 & .6313 & .0293 & .0236 & .3813 & .0521 & .0246 & .7810 \\
DA3-Giant 1.1 & \cellcolor{tabsecond}98.7 & \cellcolor{tabfirst}81.1 & \cellcolor{tabthird}63.0 & \cellcolor{tabfirst}100.0 & \cellcolor{tabsecond}96.7 & \cellcolor{tabsecond}88.2 & \cellcolor{tabfirst}97.7 & \cellcolor{tabfirst}89.2 & \cellcolor{tabsecond}70.3 & \cellcolor{tabfirst}99.4 & \cellcolor{tabfirst}98.2 & \cellcolor{tabfirst}91.7 & .1046 & .0490 & .4341 & \cellcolor{tabthird}.0109 & .0087 & \cellcolor{tabthird}.3066 & .0321 & .0131 & .3660 \\
Pi3X & \cellcolor{tabfirst}98.8 & \cellcolor{tabfirst}81.1 & \cellcolor{tabfirst}64.8 & \cellcolor{tabfirst}100.0 & 96.1 & \cellcolor{tabthird}88.1 & \cellcolor{tabsecond}97.3 & 86.7 & 66.1 & \cellcolor{tabthird}99.2 & \cellcolor{tabsecond}97.5 & 90.1 & .1255 & .0427 & .2837 & \cellcolor{tabsecond}.0107 & \cellcolor{tabfirst}.0081 & \cellcolor{tabsecond}.3059 & \cellcolor{tabsecond}.0299 & .0125 & \cellcolor{tabthird}.3467 \\
VGGT-$\Omega$ & \cellcolor{tabsecond}98.7 & \cellcolor{tabthird}80.9 & 61.7 & \cellcolor{tabfirst}100.0 & 96.4 & 88.0 & 96.7 & \cellcolor{tabsecond}87.3 & \cellcolor{tabfirst}70.6 & \cellcolor{tabsecond}99.3 & 97.3 & \cellcolor{tabthird}90.7 & .0922 & \cellcolor{tabthird}.0371 & \cellcolor{tabfirst}.2379 & .0110 & .0095 & .3123 & .0318 & \cellcolor{tabthird}.0124 & .3516 \\
DVLT & 98.4 & 80.2 & 59.7 & \cellcolor{tabfirst}100.0 & \cellcolor{tabfirst}96.9 & 87.9 & 78.6 & 67.0 & 40.4 & 99.1 & 96.7 & 81.5 & .1156 & .0428 & .3645 & \cellcolor{tabfirst}.0095 & \cellcolor{tabsecond}.0083 & .3165 & \cellcolor{tabfirst}.0280 & \cellcolor{tabfirst}.0117 & .3491 \\
ARROW & 98.4 & 78.6 & 61.6 & \cellcolor{tabfirst}100.0 & 96.0 & 87.2 & 91.1 & 85.3 & 55.6 & \cellcolor{tabthird}99.2 & 97.3 & 87.8 & \cellcolor{tabsecond}.0621 & \cellcolor{tabsecond}.0329 & .4429 & .0129 & .0101 & .3244 & .0371 & .0158 & .5023 \\
Track4World & \cellcolor{tabfirst}98.8 & 76.8 & 56.6 & \cellcolor{tabfirst}100.0 & 95.6 & 85.6 & 82.1 & 77.5 & 38.5 & \cellcolor{tabthird}99.2 & 96.9 & 83.5 & .1441 & .0658 & .3076 & .0120 & .0096 & .3092 & .0347 & .0134 & .4359 \\
\midrule
\nickname{} & \cellcolor{tabthird}98.5 & \cellcolor{tabsecond}81.0 & \cellcolor{tabsecond}63.8 & \cellcolor{tabfirst}100.0 & \cellcolor{tabthird}96.5 & \cellcolor{tabfirst}88.3 & \cellcolor{tabthird}97.1 & \cellcolor{tabthird}87.1 & \cellcolor{tabsecond}70.3 & \cellcolor{tabthird}99.2 & \cellcolor{tabthird}97.4 & \cellcolor{tabsecond}90.8 & \cellcolor{tabfirst}.0318 & \cellcolor{tabfirst}.0166 & \cellcolor{tabsecond}.2527 & .0127 & \cellcolor{tabthird}.0085 & \cellcolor{tabfirst}.3026 & .0316 & \cellcolor{tabsecond}.0119 & \cellcolor{tabfirst}.3440 \\
\bottomrule
\end{tabular*}
\par\vspace{1.5pt}
\begin{minipage}[t]{0.665\linewidth}
\fontsize{6.4}{7.2}\selectfont
\begin{tabular*}{\linewidth}[t]{@{\extracolsep{\fill}}l*{12}{c}@{}}
\toprule
\multicolumn{13}{@{}l}{\color{geomaccent}\bfseries \hypertarget{geometry-mv-recon}{(d) Multiview reconstruction (mean)}} \\
 & \multicolumn{3}{c}{\textbf{7-Scenes (dense)}} & \multicolumn{3}{c}{\textbf{NRGBD (dense)}} & \multicolumn{3}{c}{\textbf{DTU (mm)}} & \multicolumn{3}{c}{\textbf{ETH3D}} \\
\cmidrule(lr){2-4}\cmidrule(lr){5-7}\cmidrule(lr){8-10}\cmidrule(lr){11-13}
Method & Acc.$\downarrow$ & Comp.$\downarrow$ & NC$\uparrow$ & Acc.$\downarrow$ & Comp.$\downarrow$ & NC$\uparrow$ & Acc.$\downarrow$ & Comp.$\downarrow$ & NC$\uparrow$ & Acc.$\downarrow$ & Comp.$\downarrow$ & NC$\uparrow$ \\
\midrule
VGGT & .0222 & .0263 & .666 & .0147 & .0153 & .871 & 1.3095 & 1.9252 & .665 & .2713 & .3038 & .842 \\
$\pi^3$ & .0158 & \cellcolor{tabthird}.0220 & .689 & \cellcolor{tabsecond}.0128 & \cellcolor{tabfirst}.0136 & .874 & 1.1521 & 1.8006 & .668 & .1871 & .2104 & .872 \\
MapAnything & .0218 & .0233 & .671 & .0337 & .0242 & .834 & 2.1350 & 1.9010 & \cellcolor{tabfirst}.679 & .4600 & .5668 & .788 \\
DA3-Giant 1.1 & .0184 & .0232 & .687 & \cellcolor{tabthird}.0129 & \cellcolor{tabsecond}.0141 & \cellcolor{tabthird}.885 & 1.3412 & \cellcolor{tabfirst}1.6118 & \cellcolor{tabthird}.676 & .3063 & .4417 & .829 \\
Pi3X & \cellcolor{tabsecond}.0153 & .0227 & \cellcolor{tabthird}.692 & \cellcolor{tabfirst}.0120 & \cellcolor{tabthird}.0143 & .883 & 1.2268 & 1.9071 & .672 & \cellcolor{tabthird}.1811 & \cellcolor{tabsecond}.2054 & \cellcolor{tabthird}.886 \\
VGGT-$\Omega$ & .0299 & .0332 & .669 & .0160 & .0153 & .868 & \cellcolor{tabthird}1.0953 & 2.1783 & .663 & \cellcolor{tabsecond}.1733 & \cellcolor{tabfirst}.2014 & \cellcolor{tabsecond}.892 \\
DVLT & \cellcolor{tabfirst}.0149 & \cellcolor{tabfirst}.0211 & .686 & .0207 & .0175 & .866 & 1.3707 & 2.0522 & \cellcolor{tabsecond}.677 & .2575 & .2577 & .859 \\
ARROW & .0160 & .0242 & .688 & .0136 & .0146 & \cellcolor{tabsecond}.892 & \cellcolor{tabsecond}1.0613 & \cellcolor{tabsecond}1.6353 & .669 & .2719 & .3326 & .832 \\
Track4World & \cellcolor{tabthird}.0157 & \cellcolor{tabsecond}.0213 & \cellcolor{tabsecond}.696 & .0548 & .0276 & .816 & 6.2844 & 44.2710 & .569 & .3143 & .3319 & .815 \\
\midrule
\nickname{} & .0158 & .0232 & \cellcolor{tabfirst}.697 & .0149 & .0172 & \cellcolor{tabfirst}.894 & \cellcolor{tabfirst}1.0452 & \cellcolor{tabthird}1.7915 & .674 & \cellcolor{tabfirst}.1722 & \cellcolor{tabthird}.2067 & \cellcolor{tabfirst}.904 \\
\bottomrule
\end{tabular*}
\par\vspace{1.5pt}
\end{minipage}\hfill
\begin{minipage}[t]{0.31\linewidth}
\fontsize{6.4}{7.8}\selectfont
\begin{tabular*}{\linewidth}[t]{@{\extracolsep{\fill}}l*{4}{c}@{}}
\toprule
\multicolumn{5}{@{}l}{\color{geomaccent}\bfseries Metric scale (\%) $\downarrow$} \\
Method & 7-Sc. & ETH3D & NRGBD & Mean \\
\midrule
MapAnything & \cellcolor{tabthird}33.54 & \cellcolor{tabfirst}19.03 & \cellcolor{tabsecond}22.73 & \cellcolor{tabthird}25.10 \\
Pi3X & \cellcolor{tabfirst}5.41 & \cellcolor{tabthird}26.98 & \cellcolor{tabfirst}18.37 & \cellcolor{tabfirst}16.92 \\
\midrule
\nickname{} & \cellcolor{tabsecond}11.05 & \cellcolor{tabsecond}24.83 & \cellcolor{tabthird}24.44 & \cellcolor{tabsecond}20.11 \\
\bottomrule
\end{tabular*}
\par\vspace{4pt}
\raggedright\fontsize{6.4}{8}\selectfont
Scale is evaluated separately: point-cloud radius relative error, with no scale fitting or ICP; Mean equally weights datasets.
\end{minipage}
\par\vspace{2pt}
\parbox{\linewidth}{\fontsize{6.3}{7.7}\selectfont
Rank colors throughout the evaluation tables: \colorbox{tabfirst}{1st}, \colorbox{tabsecond}{2nd}, \colorbox{tabthird}{3rd}; rounded ties share color. Depth: Rel/SqRel = absolute/squared relative error, RMS/log = RMSE/log-RMSE; $\delta_i$: $\delta<1.25^i$ (\%). Pose: R/T/A = rotation/translation accuracy/AUC at the indicated angle (\%). RPE$_t$/RPE$_r$: translational/rotational relative pose error (rotation in degrees). Reconstruction: distances in m except DTU (mm); NC: normal consistency. Params.: all loaded parameters, including frozen modules; Track4World includes its tracking heads and DA3 metric branch; \nickname{}: geometry model. VGGT-$\Omega$: \texttt{VGGT-Omega-1B-416-Reproduction}. Track4World: DA3 feed-forward, RGB only, long side 518, no test-time BA.}
\fontsize{7.2}{8.7}\selectfont\setlength{\abovecaptionskip}{4pt}
\caption{\fontsize{7.2}{8.7}\selectfont\textbf{Geometry evaluation across 16 task--dataset settings.} Summary scores average unrounded metric ranks within each panel, then weight the four panels equally; pose and trajectory form one panel.}
\label{tab:geometry-results}
\endgroup
\end{table}

\paragraph{Long-sequence reconstruction.}
\label{sec:long-sequence-reconstruction}

We evaluate long-sequence reconstruction on six Oxford Spires~\citep{tao2025spires} routes with 361--800 frames per route, using the ATE, RRE, RTE, CD, and F1 metric definitions of Scal3R~\citep{xie2026scal3r}. All input images of each route are processed in a single forward pass.

As shown in Table~\ref{tab:long-sequence-results}, \nickname{} achieves the best reported results on all five metrics, reducing ATE by 47.2\% and CD by 24.1\% relative to DVLT.

\begin{table}[!htbp]
  \centering
  \begingroup
  \definecolor{tabfirst}{rgb}{1,0.7,0.7}
  \definecolor{tabsecond}{rgb}{1,0.85,0.7}
  \definecolor{tabthird}{rgb}{1,1,0.7}
  \colorlet{tabfirst}{tabfirst!55!white}
  \colorlet{tabsecond}{tabsecond!55!white}
  \colorlet{tabthird}{tabthird!55!white}
  \fontsize{8}{9.6}\selectfont\setlength{\tabcolsep}{2.5pt}
  \begin{tabular*}{\linewidth}{@{\extracolsep{\fill}}lr*{5}{r}@{}}
    \toprule
    Method & Complete & ATE$\downarrow$ & RRE$\downarrow$ & RTE$\downarrow$ & CD$\downarrow$ & F1$\uparrow$ \\
    & routes & m & $^\circ$/100\,m & m/100\,m & m & @4\,m \\
    \midrule
    VGGT & 0/6 & \textemdash{} & \textemdash{} & \textemdash{} & \textemdash{} & \textemdash{} \\
    $\pi^3$ & 6/6 & \cellcolor{tabthird}10.1852 & \cellcolor{tabthird}17.0832 & \cellcolor{tabthird}11.7017 & \cellcolor{tabthird}1.9774 & \cellcolor{tabthird}0.8668 \\
    MapAnything & 3/6 & \textemdash{} & \textemdash{} & \textemdash{} & \textemdash{} & \textemdash{} \\
    DA3-Giant 1.1 & 2/6 & \textemdash{} & \textemdash{} & \textemdash{} & \textemdash{} & \textemdash{} \\
    Pi3X & 6/6 & 10.3502 & 23.8340 & 12.2195 & 3.4667 & 0.7959 \\
    VGGT-$\Omega$ & 6/6 & 12.1434 & 28.0360 & 15.3298 & 2.4580 & 0.8328 \\
    DVLT & 6/6 & \cellcolor{tabsecond}1.3885 & \cellcolor{tabsecond}3.6422 & \cellcolor{tabsecond}2.3218 & \cellcolor{tabsecond}1.5238 & \cellcolor{tabsecond}0.9236 \\
    ARROW & 6/6 & 11.1650 & 19.8088 & 12.4884 & 2.2748 & 0.8458 \\
    Track4World & 2/6 & \textemdash{} & \textemdash{} & \textemdash{} & \textemdash{} & \textemdash{} \\
    \midrule
    \nickname{} & 6/6 & \cellcolor{tabfirst}0.7337 & \cellcolor{tabfirst}2.9298 & \cellcolor{tabfirst}1.6664 & \cellcolor{tabfirst}1.1573 & \cellcolor{tabfirst}0.9392 \\
    \bottomrule
  \end{tabular*}
  \par\vspace{3pt}
  \parbox{\linewidth}{\fontsize{7.2}{8.6}\selectfont
    Complete: successfully evaluated routes out of six. \textemdash{}: six-route mean unavailable due to out-of-memory (OOM) failures.
  }
  \endgroup
  \caption{\textbf{Long-sequence reconstruction on Oxford Spires.}}
  \label{tab:long-sequence-results}
\end{table}

\subsection{2D and 3D Tracking}
\label{sec:tracking-evaluation}

We follow Track4World~\citep{lu2026track4world} on its released subsets of Aria Digital Twin (ADT)~\citep{pan2023adt}, PointOdyssey~\citep{zheng2023pointodyssey}, and Panoptic Studio (PStudio)~\citep{joo2015panoptic}, with ADT and PStudio annotations from TAPVid-3D~\citep{koppula2024tapvid3d}. These datasets cover 298 clip--horizon cases, equally averaging 16- and 50-frame settings. We compare with Track4World, MoSE3~\citep{cheng2026mose3}, ARROW~\citep{fradlin2026arrow}, CoWTracker~\citep{lai2026cowtracker}, and TAPIP3D~\citep{zhang2025tapip3d}, reporting APD and AJ for 2D and world-space 3D tracking.

For the methods marked $\dagger$, CoWTracker lifts its unchanged 2D tracks using predicted depth and cameras, retaining its original visibility; TAPIP3D tracks natively in 3D and projects to 2D. All 3D results use scale-and-shift alignment fitted to ground-truth-visible points; geometry confidence does not filter evaluated predictions.

\emph{GameRealm (Long)}~\citep{gamerealm} contains 45 clips from 15 source sequences in games unseen during motion training, with 256 consecutive native frames per clip. Scores average clips within source sequence, sequences within game, and games equally. Its 3D scoring uses alignment, median rescaling, and approximate metric units. \nickname{} leads Average 2D AJ and ranks second in the other three metrics (Table~\ref{tab:tracking-results}(a)).

\subsection{Multiview Tracking}
\label{sec:multiview-tracking-evaluation}

We evaluate RGB-only multiview tracking on 58 scene--rig cases: 10 DexYCB~\citep{chao2021dexycb}, 30 Kubric~\citep{greff2022kubric}, and 18 Panoptic Studio~\citep{joo2015panoptic}, with four cameras and 24 synchronized timesteps. Each time--camera observation uses all registered queries and the same geometry sources as single-view tracking. 3D scoring uses scale-and-translation alignment followed by median rescaling; Kubric units are approximate. Scores average cases within each dataset.

As shown in Table~\ref{tab:multiview-tracking-results}(b), \nickname{} leads dataset-average 2D AJ and ranks second in 2D APD and both 3D metrics.

\begin{table}[!htbp]
  \centering
  \begingroup
  \definecolor{tabfirst}{rgb}{1,0.7,0.7}
  \definecolor{tabsecond}{rgb}{1,0.85,0.7}
  \definecolor{tabthird}{rgb}{1,1,0.7}
  \colorlet{tabfirst}{tabfirst!55!white}
  \colorlet{tabsecond}{tabsecond!55!white}
  \colorlet{tabthird}{tabthird!55!white}
  \definecolor{trackaccent}{HTML}{174B68}
  \fontsize{6.6}{8.2}\selectfont
  \setlength{\tabcolsep}{0.8pt}
  \renewcommand{\arraystretch}{1}
  \setlength{\parskip}{0pt}
  \setlength{\aboverulesep}{0.7pt}
  \setlength{\belowrulesep}{0.7pt}
  \setlength{\cmidrulesep}{0pt}
  \setlength{\cmidrulekern}{1pt}
  \begin{tabular*}{\linewidth}{@{\extracolsep{\fill}}l*{4}{c}*{4}{c}*{4}{c}*{4}{c}!{\hspace{5pt}}*{4}{c}@{}}
    \toprule
    \multicolumn{21}{@{}l}{\color{trackaccent}\bfseries (a) Single-view tracking} \\
    & \multicolumn{4}{c}{\textbf{ADT}} & \multicolumn{4}{c}{\textbf{PointOdyssey}} & \multicolumn{4}{c}{\textbf{PStudio}} & \multicolumn{4}{c}{\textbf{GameRealm (Long)}} & \multicolumn{4}{c}{\textbf{Average}} \\
    \cmidrule(lr){2-5}\cmidrule(lr){6-9}\cmidrule(lr){10-13}\cmidrule(lr){14-17}\cmidrule(lr){18-21}
    & \multicolumn{2}{c}{2D} & \multicolumn{2}{c}{3D} & \multicolumn{2}{c}{2D} & \multicolumn{2}{c}{3D} & \multicolumn{2}{c}{2D} & \multicolumn{2}{c}{3D} & \multicolumn{2}{c}{2D} & \multicolumn{2}{c}{3D} & \multicolumn{2}{c}{2D} & \multicolumn{2}{c}{3D} \\
    \cmidrule(lr){2-3}\cmidrule(lr){4-5}\cmidrule(lr){6-7}\cmidrule(lr){8-9}\cmidrule(lr){10-11}\cmidrule(lr){12-13}\cmidrule(lr){14-15}\cmidrule(lr){16-17}\cmidrule(lr){18-19}\cmidrule(lr){20-21}
    Method & APD$\uparrow$ & AJ$\uparrow$ & APD$\uparrow$ & AJ$\uparrow$ & APD$\uparrow$ & AJ$\uparrow$ & APD$\uparrow$ & AJ$\uparrow$ & APD$\uparrow$ & AJ$\uparrow$ & APD$\uparrow$ & AJ$\uparrow$ & APD$\uparrow$ & AJ$\uparrow$ & APD$\uparrow$ & AJ$\uparrow$ & APD$\uparrow$ & AJ$\uparrow$ & APD$\uparrow$ & AJ$\uparrow$ \\
    \midrule
    Track4World & \cellcolor{tabthird}93.16 & \cellcolor{tabthird}88.46 & 65.78 & 60.28 & \cellcolor{tabthird}90.19 & \cellcolor{tabthird}66.90 & 52.88 & \cellcolor{tabthird}37.95 & \cellcolor{tabthird}80.49 & \cellcolor{tabthird}68.36 & 59.92 & 50.39 & 67.09\textsuperscript{*} & 50.19\textsuperscript{*} & 24.68\textsuperscript{*} & 18.22\textsuperscript{*} & 82.73\textsuperscript{*} & 68.48\textsuperscript{*} & 50.81\textsuperscript{*} & 41.71\textsuperscript{*} \\
    MoSE3 & 83.17 & 76.96 & 64.64 & 59.07 & 82.84 & 61.11 & \cellcolor{tabfirst}56.55 & \cellcolor{tabfirst}41.59 & 75.67 & 64.51 & \cellcolor{tabsecond}66.10 & \cellcolor{tabfirst}57.42 & 39.22 & 27.08 & \cellcolor{tabsecond}38.59 & \cellcolor{tabsecond}29.82 & 70.23 & 57.41 & \cellcolor{tabthird}56.47 & \cellcolor{tabthird}46.97 \\
    ARROW & 80.60 & 73.59 & \cellcolor{tabfirst}77.95 & \cellcolor{tabfirst}71.76 & 85.38 & 62.06 & \cellcolor{tabsecond}56.21 & \cellcolor{tabsecond}41.10 & 76.96 & 64.01 & \cellcolor{tabfirst}67.15 & \cellcolor{tabsecond}57.14 & 16.79 & 10.09 & 33.78 & 24.44 & 64.93 & 52.44 & \cellcolor{tabfirst}58.77 & \cellcolor{tabfirst}48.61 \\
    CoWTracker\textsuperscript{\textdagger} & \cellcolor{tabfirst}94.56 & \cellcolor{tabfirst}90.28 & \cellcolor{tabthird}70.76 & \cellcolor{tabthird}65.05 & \cellcolor{tabfirst}91.40 & \cellcolor{tabsecond}67.76 & 50.98 & 36.74 & \cellcolor{tabfirst}84.39 & \cellcolor{tabsecond}72.17 & 65.08 & \cellcolor{tabthird}55.84 & \cellcolor{tabfirst}79.33 & \cellcolor{tabsecond}59.48 & \cellcolor{tabthird}37.05 & \cellcolor{tabthird}29.06 & \cellcolor{tabfirst}87.42 & \cellcolor{tabsecond}72.42 & 55.97 & 46.68 \\
    TAPIP3D\textsuperscript{\textdagger} & 92.32 & 86.48 & 70.32 & 63.50 & 88.85 & 64.24 & \cellcolor{tabthird}53.14 & 37.57 & 78.15 & 64.13 & \cellcolor{tabthird}65.18 & 54.34 & \cellcolor{tabthird}48.28 & \cellcolor{tabthird}32.20 & 34.08 & 23.91 & \cellcolor{tabthird}76.90 & \cellcolor{tabthird}61.76 & 55.68 & 44.83 \\
    \midrule
    \nickname{} & \cellcolor{tabsecond}93.68 & \cellcolor{tabsecond}88.92 & \cellcolor{tabsecond}72.90 & \cellcolor{tabsecond}66.30 & \cellcolor{tabsecond}91.28 & \cellcolor{tabfirst}68.80 & 52.65 & 37.62 & \cellcolor{tabsecond}83.27 & \cellcolor{tabfirst}72.27 & 64.39 & 55.30 & \cellcolor{tabsecond}77.29 & \cellcolor{tabfirst}59.71 & \cellcolor{tabfirst}39.36 & \cellcolor{tabfirst}31.47 & \cellcolor{tabsecond}86.38 & \cellcolor{tabfirst}72.43 & \cellcolor{tabsecond}57.33 & \cellcolor{tabsecond}47.68 \\
    \bottomrule
  \end{tabular*}
  \par\vspace{4pt}
  \begin{tabular*}{\linewidth}{@{\extracolsep{\fill}}l*{4}{c}*{4}{c}*{4}{c}!{\hspace{5pt}}*{4}{c}@{}}
    \toprule
    \multicolumn{17}{@{}l}{\color{trackaccent}\bfseries (b) Multiview tracking} \\
    & \multicolumn{4}{c}{\textbf{DexYCB}} & \multicolumn{4}{c}{\textbf{Kubric}} & \multicolumn{4}{c}{\textbf{Panoptic}} & \multicolumn{4}{c}{\textbf{Average}} \\
    \cmidrule(lr){2-5}\cmidrule(lr){6-9}\cmidrule(lr){10-13}\cmidrule(lr){14-17}
    & \multicolumn{2}{c}{2D} & \multicolumn{2}{c}{3D} & \multicolumn{2}{c}{2D} & \multicolumn{2}{c}{3D} & \multicolumn{2}{c}{2D} & \multicolumn{2}{c}{3D} & \multicolumn{2}{c}{2D} & \multicolumn{2}{c}{3D} \\
    \cmidrule(lr){2-3}\cmidrule(lr){4-5}\cmidrule(lr){6-7}\cmidrule(lr){8-9}\cmidrule(lr){10-11}\cmidrule(lr){12-13}\cmidrule(lr){14-15}\cmidrule(lr){16-17}
    Method & APD$\uparrow$ & AJ$\uparrow$ & APD$\uparrow$ & AJ$\uparrow$ & APD$\uparrow$ & AJ$\uparrow$ & APD$\uparrow$ & AJ$\uparrow$ & APD$\uparrow$ & AJ$\uparrow$ & APD$\uparrow$ & AJ$\uparrow$ & APD$\uparrow$ & AJ$\uparrow$ & APD$\uparrow$ & AJ$\uparrow$ \\
    \midrule
    Track4World & \cellcolor{tabthird}49.61 & \cellcolor{tabsecond}35.46 & 51.39 & 25.77 & 35.35 & \cellcolor{tabthird}23.04 & 37.44 & 11.82 & \cellcolor{tabthird}58.21 & 34.71 & 39.60 & 22.17 & \cellcolor{tabthird}47.72 & \cellcolor{tabthird}31.07 & 42.81 & 19.92 \\
    MoSE3 & 46.44 & 23.43 & 59.28 & \cellcolor{tabthird}31.44 & 32.21 & 17.15 & 54.87 & \cellcolor{tabthird}31.01 & 57.78 & \cellcolor{tabsecond}37.94 & 46.49 & \cellcolor{tabsecond}31.42 & 45.48 & 26.17 & 53.55 & \cellcolor{tabthird}31.29 \\
    ARROW & 40.06 & 17.38 & \cellcolor{tabsecond}70.36 & \cellcolor{tabsecond}33.51 & \cellcolor{tabthird}38.69 & 22.02 & \cellcolor{tabfirst}69.40 & \cellcolor{tabfirst}43.93 & \cellcolor{tabfirst}63.09 & \cellcolor{tabfirst}40.79 & \cellcolor{tabfirst}70.25 & \cellcolor{tabfirst}48.09 & 47.28 & 26.73 & \cellcolor{tabfirst}70.00 & \cellcolor{tabfirst}41.84 \\
    CoWTracker\textsuperscript{\textdagger} & 33.68 & 17.56 & 49.24 & 15.25 & 21.10 & 11.88 & 35.89 & 9.93 & 30.82 & 14.22 & 38.07 & 17.64 & 28.53 & 14.55 & 41.07 & 14.27 \\
    TAPIP3D\textsuperscript{\textdagger} & \cellcolor{tabsecond}57.75 & \cellcolor{tabfirst}35.69 & \cellcolor{tabthird}65.41 & 31.30 & \cellcolor{tabfirst}50.36 & \cellcolor{tabsecond}26.33 & \cellcolor{tabthird}64.72 & 27.36 & \cellcolor{tabsecond}62.99 & 34.95 & \cellcolor{tabsecond}55.61 & 28.88 & \cellcolor{tabfirst}57.03 & \cellcolor{tabsecond}32.32 & \cellcolor{tabthird}61.91 & 29.18 \\
    \midrule
    \nickname{} & \cellcolor{tabfirst}61.11 & \cellcolor{tabthird}35.22 & \cellcolor{tabfirst}74.42 & \cellcolor{tabfirst}37.86 & \cellcolor{tabsecond}49.95 & \cellcolor{tabfirst}32.10 & \cellcolor{tabsecond}65.67 & \cellcolor{tabsecond}32.25 & 58.18 & \cellcolor{tabthird}37.90 & \cellcolor{tabthird}54.96 & \cellcolor{tabthird}30.99 & \cellcolor{tabsecond}56.41 & \cellcolor{tabfirst}35.07 & \cellcolor{tabsecond}65.01 & \cellcolor{tabsecond}33.70 \\
    \bottomrule
  \end{tabular*}
  \par\vspace{3pt}
  \parbox{\linewidth}{\fontsize{6.3}{7.7}\selectfont
    \textsuperscript{*}Track4World scores 37/45 GameRealm (Long) clips; eight clips with nonfinite outputs are excluded. Other methods score 45/45. Its long-sequence entries and resulting Average have different coverage and are not ranked.
    \textsuperscript{\textdagger}CoWTracker and TAPIP3D use VGGT-$\Omega$~\citep{wang2026vggtomega} geometry with separate tracking models; other methods jointly model reconstruction and tracking.
  }
  \fontsize{7.2}{8.7}\selectfont\setlength{\abovecaptionskip}{4pt}
  \caption{\fontsize{7.2}{8.7}\selectfont\textbf{Single-view (a) and multiview (b) tracking.} APD and AJ are percentages (higher is better). Average equally weights the datasets in each panel.}
  \label{tab:tracking-evaluation}
  \label{tab:tracking-results}
  \label{tab:multiview-tracking-results}
  \endgroup
\end{table}

\subsection{Moving-Object Segmentation}
\label{sec:motion-mask-evaluation}

Following GMOS~\citep{xie2026gmos}, we evaluate binary moving-object segmentation on SegTrackv2, the FBMS test split, and YTVOS19-M using region similarity $\mathcal{J}$, contour accuracy $\mathcal{F}$, and their average $\mathcal{J}\&\mathcal{F}$.

As shown in Table~\ref{tab:motion-mask-results}, \nickname{} achieves the best region similarity on SegTrackv2 and the second-highest mean $\mathcal{J}\&\mathcal{F}$ of 85.04.

\begin{table}[!htbp]
\centering\begingroup
\definecolor{tabfirst}{rgb}{1,0.7,0.7}
\definecolor{tabsecond}{rgb}{1,0.85,0.7}
\definecolor{tabthird}{rgb}{1,1,0.7}
\colorlet{tabfirst}{tabfirst!55!white}
\colorlet{tabsecond}{tabsecond!55!white}
\colorlet{tabthird}{tabthird!55!white}
\fontsize{7}{8.5}\selectfont
\setlength{\tabcolsep}{1.2pt}\renewcommand{\arraystretch}{1.05}
\begin{tabular*}{\linewidth}{@{\extracolsep{\fill}}l*{12}{c}@{}}
\toprule
 & \multicolumn{3}{c}{\textbf{SegTrackv2}} & \multicolumn{3}{c}{\textbf{FBMS Test}} & \multicolumn{3}{c}{\textbf{YTVOS19-M}} & \multicolumn{3}{c}{\textbf{Mean}} \\
\cmidrule(lr){2-4}\cmidrule(lr){5-7}\cmidrule(lr){8-10}\cmidrule(lr){11-13}
Method & $\mathcal{J}$ & $\mathcal{F}$ & $\mathcal{J}\&\mathcal{F}$ & $\mathcal{J}$ & $\mathcal{F}$ & $\mathcal{J}\&\mathcal{F}$ & $\mathcal{J}$ & $\mathcal{F}$ & $\mathcal{J}\&\mathcal{F}$ & $\mathcal{J}$ & $\mathcal{F}$ & $\mathcal{J}\&\mathcal{F}$ \\
\midrule
FlowSAM & 77.90 & 83.90 & 80.90 & 82.80 & 80.90 & 81.85 & 52.40 & 54.30 & 53.35 & 71.03 & 73.03 & 72.03 \\
SegAnyMo & 76.30 & 85.40 & 80.85 & 78.30 & 82.80 & 80.55 & 70.90 & 73.00 & 71.95 & 75.17 & 80.40 & 77.78 \\
GeoMotion & 77.30 & 84.30 & 80.80 & 72.50 & 78.50 & 75.50 & 65.40 & 65.70 & 65.55 & 71.73 & 76.17 & 73.95 \\
GMOS & \cellcolor{tabsecond}82.00 & \cellcolor{tabfirst}88.30 & \cellcolor{tabfirst}85.15 & \cellcolor{tabfirst}91.90 & \cellcolor{tabfirst}93.00 & \cellcolor{tabfirst}92.45 & \cellcolor{tabfirst}85.10 & \cellcolor{tabfirst}84.60 & \cellcolor{tabfirst}84.85 & \cellcolor{tabfirst}86.33 & \cellcolor{tabfirst}88.63 & \cellcolor{tabfirst}87.48 \\
GMOS-S & \cellcolor{tabthird}78.90 & \cellcolor{tabthird}85.90 & \cellcolor{tabthird}82.40 & \cellcolor{tabthird}83.00 & \cellcolor{tabthird}84.90 & \cellcolor{tabthird}83.95 & \cellcolor{tabthird}81.40 & \cellcolor{tabthird}81.80 & \cellcolor{tabthird}81.60 & \cellcolor{tabthird}81.10 & \cellcolor{tabthird}84.20 & \cellcolor{tabthird}82.65 \\
\midrule
\nickname{} & \cellcolor{tabfirst}83.42 & \cellcolor{tabsecond}86.19 & \cellcolor{tabsecond}84.80 & \cellcolor{tabsecond}86.42 & \cellcolor{tabsecond}86.56 & \cellcolor{tabsecond}86.49 & \cellcolor{tabsecond}84.88 & \cellcolor{tabsecond}82.78 & \cellcolor{tabsecond}83.83 & \cellcolor{tabsecond}84.91 & \cellcolor{tabsecond}85.18 & \cellcolor{tabsecond}85.04 \\
\bottomrule\end{tabular*}
\caption{\fontsize{7.2}{8.7}\selectfont\textbf{Moving-object segmentation.} Scores are percentages (higher is better); Mean equally weights the three datasets. Baseline results are from GMOS Table~2~\citep{xie2026gmos}, with means computed from its published rounded scores.}
\label{tab:motion-mask-results}
\endgroup\end{table}

\subsection{Qualitative Results}
\label{sec:qualitative-generalization}

\begin{figure}[!htbp]
  \centering
  \includegraphics[width=\linewidth]{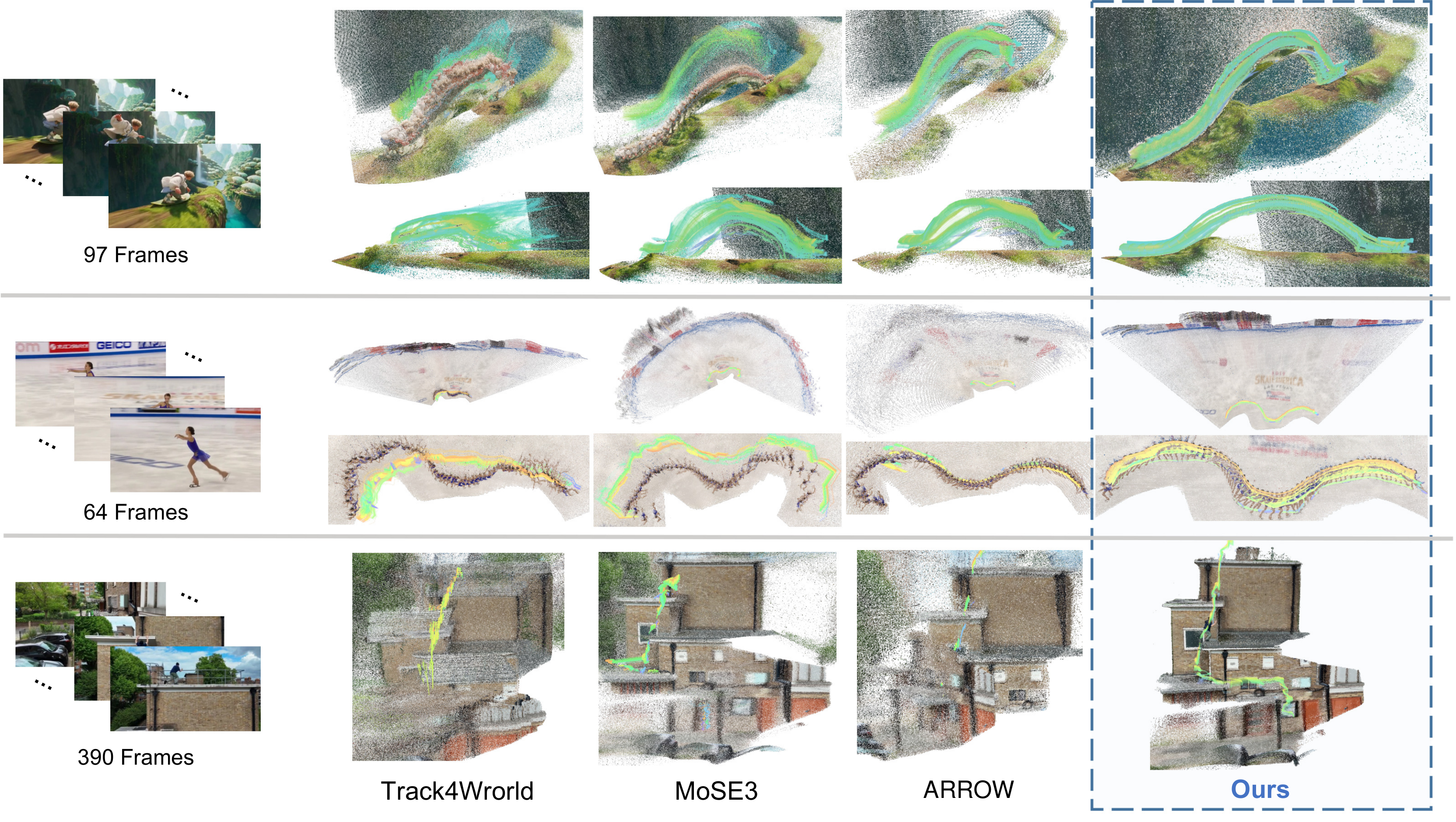}
  \caption{\textbf{Qualitative 3D tracking comparison.} Left: input frames. Remaining columns: reconstructed scenes and colored 3D trajectories from Track4World, MoSE3, ARROW, and \nickname{}. From top to bottom: 97, 64, and 390 frames.}
  \label{fig:tracking-comparison}
\end{figure}

Figures~\ref{fig:teaser}, \ref{fig:reconstruction-comparison}, and~\ref{fig:tracking-comparison} show reconstructed geometry and motion across scenes and visual styles. Figure~\ref{fig:tracking-comparison} illustrates the geometric consistency of the lifted tracks. In the 64-frame example (second row), ARROW's trajectories at later frames with cameras farther from the first-frame origin deviate visibly from the geometry reconstructed for the corresponding frames, whereas our lifted trajectories remain registered with the reconstructed surfaces. This observation is consistent with the error coupling that arises when independently regressed 3D endpoints are composed with camera estimates in a first-frame reference system.

\FloatBarrier
\section{Design Choices and Empirical Observations}
\label{sec:design-lessons}

We study the architectural and training choices underlying \nickname{} through targeted ablations and qualitative comparisons, highlighting the empirical findings and practical trade-offs that inform our final design. For loss-based ablations, we use matched model-development configurations, and percentage changes refer to relative differences in validation loss unless otherwise stated.

\subsection{Looped Transformers: Training Dynamics and Step Conditioning}
\label{sec:loop-lessons}

\begin{figure}[!htbp]
  \centering
  \includegraphics[width=\linewidth]{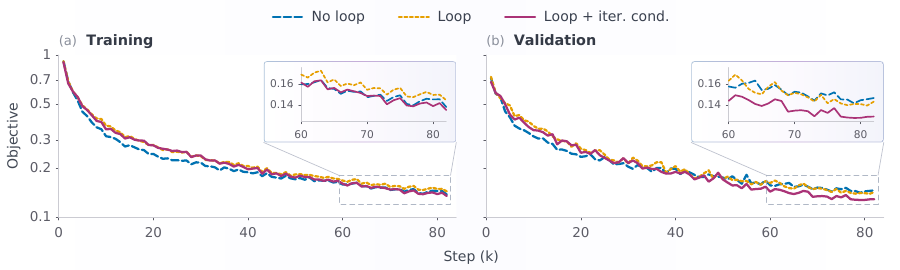}
  \caption{\textbf{Training dynamics of looped decoders.} The iteration-conditioned looped decoder achieves lower training and validation objectives than the non-looped baseline later in training.}
  \label{fig:loop-training-curves}
\end{figure}

Following DVLT~\citep{burzio2026dejaview}, we use a looped Transformer decoder with explicit loop-step conditioning. The looped and non-looped reconstruction models follow similar early training trajectories, but the looped model's advantage emerges later in training and is more pronounced with step conditioning. Figure~\ref{fig:loop-training-curves} shows the training and validation objectives of these decoder variants. A separate comparison within the looped architecture isolates the effect of conditioning: adding the step signal reduces local point loss by 2.8\% and camera loss by 10.2\% relative to the unconditioned baseline. These results support the use of step conditioning in our decoder.

\FloatBarrier
\subsection{Training View Counts and Memory Allocation}
\label{sec:training-view-counts}

\begin{wrapfigure}{r}{0.49\textwidth}
  \centering
  \includegraphics[width=\linewidth]{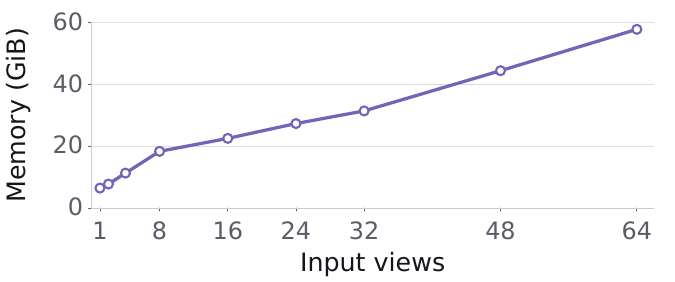}
  \caption{\textbf{Training memory.} Peak allocated memory per full update on an A800 80GB GPU: $512\times512$ inputs, BF16, 16 loop steps, and activation recomputation.}
  \label{fig:training-memory}
\end{wrapfigure}

We allocate part of the training memory budget to larger joint-view contexts, rather than using it only to pack more independent short examples. Our training recipe spans single-image inputs, short view sets of 2--24 images, and long view sets of 32, 48, or 64 images. View counts are sampled stochastically: short view sets receive most of the probability mass, with additional mass assigned to single-image inputs and the long-view buckets. This mixture combines frequent small-context updates with regular exposure to larger sets of observations.

The view count refers to images from one scene or sequence, not the aggregate image count across independent examples in a batch. Including larger view sets gives the model direct training exposure to combining evidence across many observations, rather than relying exclusively on extrapolation from short inputs. This training choice complements weight sharing in the looped architecture. Figure~\ref{fig:training-memory} reports the measured training-memory footprint for input sets up to 64 views.

\subsection{Encoder Layer Selection and Fine-Tuning}
\label{sec:feature-lessons}

\paragraph{Layer selection.}
UFM~\citep[Appendix~E]{zhang2025ufm} shows that intermediate features from a frozen encoder can outperform its final-layer features for dense prediction. Our geometry ablations using DINOv3~\citep{simeoni2025dinov3} show a consistent trend. Jointly using layers 11 and 17 gives the best local point reconstruction and camera estimation among the combinations evaluated. This pair, also used in RoMa~v2~\citep{edstedt2025romav2}, supplies the input to our multi-view decoder.

\paragraph{Freezing and learning rate.}
Fine-tuning the encoder at $10^{-4}$ initially reduces the loss faster than freezing the encoder, but performs worse later in training. By contrast, fine-tuning at $10^{-5}$ reduces local point loss by 8.1\% and camera loss by 4.5\% relative to freezing. The comparison therefore depends on both learning rate and training duration. Our encoder remains trainable, with a lower learning rate than the decoder and prediction heads.

\subsection{Output Factorization and Geometric Consistency}
\label{sec:factorization-lessons}

\paragraph{Local geometry and camera placement.}
Following $\pi^3$~\citep{wang2025pi3}, we predict camera-local pointmaps and camera poses rather than directly regressing a global pointmap. Global point errors mix errors in local surface geometry with errors in its cross-view placement. Image evidence may constrain local surface structure even when the relative camera configuration remains uncertain. Separate local-point and camera supervision preserves this distinction, allowing local geometry to be learned without requiring global placement to be equally accurate.

\paragraph{Correspondence and 3D tracking.}
Factorizing 3D tracking into image-space correspondence prediction and geometric lifting avoids asking the motion head to independently reproduce target-view geometry and camera placement when predicting 3D endpoints or displacements. For visible tracks, lifting enforces agreement with the target geometry at the predicted correspondence by construction; independently predicted 3D positions need not satisfy this constraint.

\begin{wrapfigure}{r}{0.5\textwidth}
  \centering
  \includegraphics[width=\linewidth]{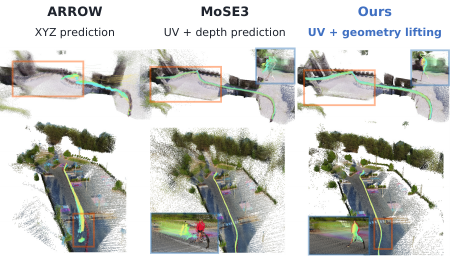}
  \caption{\textbf{Motion parameterization.} Top: local track--geometry close-ups; bottom: global tracks. ARROW regresses XYZ, MoSE3 predicts UV and depth, and \nickname{} lifts UV through geometry. Independent depth can separate MoSE3 tracks from reconstructed surfaces; ours remain registered by construction.}
  \label{fig:motion-parameterization}
\end{wrapfigure}

Figure~\ref{fig:motion-parameterization} compares each method's native 3D tracks with its own reconstruction. In the global views, ARROW's directly regressed reference-frame endpoints degrade as target cameras move farther from the query view and the scene extent grows; late trajectories drift far from the corresponding scene structure. MoSE3 and \nickname{} produce similarly coherent global trajectories, consistent with the numerical benefit of regressing image-space coordinates rather than large reference-frame XYZ values.

The local close-ups reveal a different source of error. MoSE3 lifts its UV predictions using an independently predicted tracking depth, which can place a track off the reconstructed surface even when its global trajectory looks plausible. \nickname{} instead samples target-view geometry at the predicted correspondence, without a separate XYZ or tracking-depth prediction. Visible endpoints therefore lie on the reconstructed surface by construction, removing this independent source of track--geometry disagreement.

\subsection{Geometry-Conditioned Motion}
\label{sec:geometry-motion-lessons}

Our geometry-conditioned motion branch is designed to reuse the spatial correspondence cues integrated by the geometry branch across views. Rather than independently learning cross-view spatial relationships, the motion branch conditions on this representation through Spatial AdaLN. In our ablation, geometry-feature conditioning reduces EPE by 7.17~px overall and 8.10~px on large-camera-baseline pairs compared with an otherwise identical model without this conditioning. These improvements show that geometry features provide useful spatial context for motion estimation, particularly for image pairs with large camera baselines.

\subsection{Camera Supervision and Gradient Routing}
\label{sec:camera-lessons}

\paragraph{Camera targets.}
Our camera ablations favor the reference-free formulation of $\pi^3$ with relative pose supervision~\citep{wang2025pi3} over regression tied to an explicitly designated reference frame, with more efficient convergence. We therefore use relative pose supervision, together with patch-matching supervision~\citep{wang2026vggtomega} to support learning at initialization, when relative-pose optimization is difficult~\citep[Appendix~A.4]{wang2025pi3}.

\paragraph{Camera readout.}
In an early parameter-matched comparison, patch-token readout gives lower local point and camera losses than readout from a dedicated register token. We therefore retain patch-token readout for the camera decoder.

\paragraph{Normalization gradients.}
Camera translation shares the predicted local-point normalization defined in Section~\ref{sec:method-geometry-loss}. In that section's notation, $s(\mathbf{P})$ is the sequence-wide mean valid norm of predicted local points, distinct from the subsequently fitted alignment scale $a$. We normalize the points while stopping the gradient through this same $s(\mathbf{P})$ on the camera-translation path:
\begin{equation}
  \widetilde{\mathbf{P}}=\frac{\mathbf{P}}{s(\mathbf{P})},
  \qquad
  \widetilde{\mathbf{t}}=\frac{\mathbf{t}}{\stopgrad\!\left(s(\mathbf{P})\right)}.
  \label{eq:camera-scale-stopgrad}
\end{equation}
This preserves the shared normalization in the forward pass but prevents the camera objective from changing the point normalization through this gradient path. The local-point objective still differentiates through $s(\mathbf{P})$. The operation neither detaches the camera prediction nor removes interactions between tasks through their shared representation.

In the normalization ablation, detaching this gradient path reduces camera loss by 10.9\%, while local point loss remains nearly unchanged, with a 0.2\% increase. Our camera objective uses this gradient-routing choice.

\subsection{Relative-Point Supervision and Loss Weighting}
\label{sec:loss-lessons}

Supervising geometry constructed from multiple predicted quantities is established practice: VGGT-$\Omega$~\citep{wang2026vggtomega} forms point maps from depth and camera predictions, while DA3~\citep{chen2025depthanything3} supervises points constructed from depth and rays. We use relative-point supervision alongside local-point and camera objectives. In the loss-weight ablation, increasing the relative-point weight by an order of magnitude reduces local point loss by 5.4\% and camera loss by 2.3\%. Removing this term and compensating with a stronger local-point objective does not recover the same camera quality; further increasing the local-point weight does not consistently help. These comparisons guide the loss weighting in our model.

\subsection{Dense Prediction Heads: Practical Trade-offs}
\label{sec:head-lessons}

Motivated by PXDepth's monocular-depth results, we tested a PXDepth/PixelDiT-inspired pixel-level head~\citep{yuan2026pxdepth,yu2025pixeldit} in place of the standard convolutional head. Patch-aligned artifacts persisted in local point predictions, particularly on high-uncertainty inputs. VGGT-$\Omega$ reports related blocky artifacts for all-MLP depth decoding in uncertain regions~\citep[Section~3.1.3]{wang2026vggtomega}.

We therefore retain the standard convolutional head for local point prediction. It reads the final decoder representation, without additional multi-layer feature fusion, to keep the design simple.

\subsection{Surface Structure Refinement with 3D Sparse Convolution}
\label{sec:refinement-lessons}

Following MoGe3~\citep{kong2026moge3}, we include a 3D sparse-convolutional refinement stage that aggregates over geometric neighborhoods, reducing mixing between image-adjacent but depth-separated surfaces.

\begin{figure}[!htbp]
  \centering
  \includegraphics[width=\linewidth]{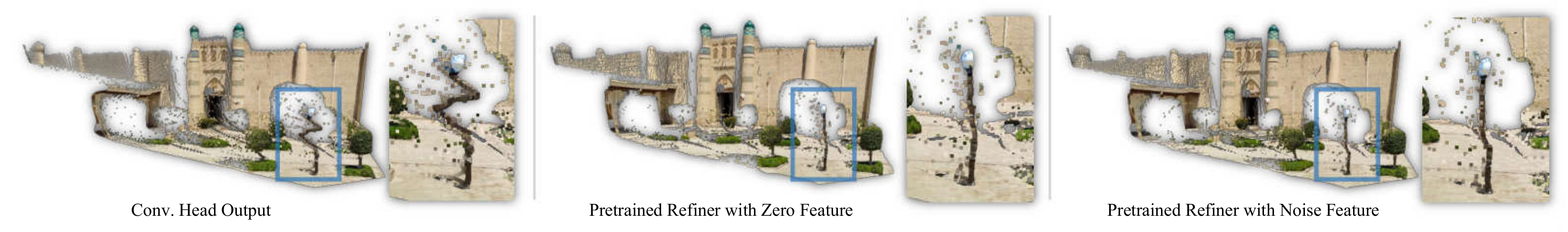}
  \caption{\textbf{Surface refinement without informative image conditioning.} From left to right: convolutional-head output, and pretrained-refiner results with image-derived conditioning features replaced by zeros or random noise, respectively. Both settings improve local structure while increasing point-placement error. Each comparison pairs an overview with a close-up of the blue-boxed thin structure.}
  \label{fig:refiner-conditioning}
\end{figure}

In qualitative zero-shot tests, we apply the pretrained MoGe3 refiner to point clouds predicted by the convolutional head, replacing its image-derived conditioning features with zeros or random noise. In the examples shown in Figure~\ref{fig:refiner-conditioning}, both settings yield more coherent surfaces and reduce distortions in thin structures near depth discontinuities.

Despite these structural improvements, zero and random image conditioning increase mean normalized XYZ $L_1$ error by 8.77\% and 7.47\%, respectively, relative to the coarse prediction in our diagnostic. We therefore condition our refiner on projected geometry-decoder features to guide structural corrections toward image-supported surface locations.

\FloatBarrier

\FloatBarrier
\section{Limitations}
\label{sec:limitations}

\paragraph{Camera--object motion ambiguity.}
Although foreground-suppressed pseudo-labeling (Section~\ref{sec:internet-video-curation}) reduces reliance on dynamic foreground cues, \nickname{} can still interpret a moving foreground as stationary when the static background is weakly textured or the foreground is visually dominant. In these cases, the model confounds camera motion with object motion, yielding a reconstruction that is inconsistent with the static background. Resolving this ambiguity requires scene and motion understanding beyond local texture correspondences.

\paragraph{Sensitivity to distance from the coordinate origin.}
We observe larger camera-position errors for cameras farther from the reconstruction origin than for those near it. This suggests that pose accuracy may be sensitive to the coordinate representation and the numerical scale of absolute translations. Understanding this dependence and reducing sensitivity to the choice of coordinate origin and translation magnitude remain open problems.

\paragraph{Accuracy and long-sequence generalization.}
\nickname{} does not consistently match the accuracy of traditional geometric optimization methods, particularly when inference sequences are longer than those seen during training. Maintaining globally consistent camera trajectories and reconstructions under such sequence-length extrapolation remains challenging. Improving long-sequence generalization and closing the accuracy gap to optimization-based methods are important directions for future work.

\section{Conclusion}

We have presented \nickname{}, a unified foundation model for 4D scene perception from diverse, uncalibrated RGB observations. With a shared backbone and approximately 0.5B parameters, it combines dense geometry, camera pose, and metric-scale estimation with 2D/3D point tracking and moving-object segmentation. By predicting image-space correspondences conditioned on geometry and lifting them into the reconstructed 3D scene, \nickname{} unifies geometry and motion in a single forward pass.

Across diverse benchmarks, \nickname{} achieves the best overall average rank in geometry evaluation and competitive performance in single- and multiview 2D/3D tracking and moving-object segmentation. It generalizes to diverse in-the-wild observations, producing coherent joint reconstructions and tracks from thousands of input images in large-scale scenes. Together with our data and training recipe and empirical studies of key design choices, these results mark a step towards unified 4D scene perception.

\begingroup
\small
\setlength{\bibsep}{2pt}
\bibliographystyle{unsrtnat}
\bibliography{references}
\endgroup
\end{document}